\documentclass[nohyperref]{article}

\usepackage{microtype}
\usepackage{graphicx}
\usepackage{subfigure}
\usepackage{booktabs} %
\usepackage{natbib}
\usepackage{multirow} 
\usepackage[colorlinks,citecolor=DarkGreen,linkcolor=FireBrick,urlcolor=FireBrick]{hyperref}

\usepackage{graphicx}
\usepackage{tikz}
\usepackage{tikz-cd}
\usepackage{times}
\usepackage{courier}
\usepackage{bm}
\usepackage{physics}
\usepackage{xcolor}
\usepackage{natbib}
\usepackage{mdframed}
\usepackage{nicefrac}
\usepackage{booktabs}
\usepackage{blindtext}
\usepackage{braket}
\usepackage{amsmath}
\usepackage{amssymb}
\usepackage{mathtools}
\usepackage{titletoc}
\usepackage{enumitem}

\usepackage{cite}
\usepackage{graphicx}
\usepackage{tikz}
\usepackage{tikz-cd}
\usepackage{times}
\usepackage{courier}
\usepackage{bm}
\usepackage{physics}
\usepackage{xcolor}
\usepackage{natbib}
\usepackage{mdframed}
\usepackage{nicefrac}
\usepackage{booktabs}
\usepackage{lipsum}
\usepackage{titlesec}
\usepackage{wrapfig,lipsum,booktabs}
\usepackage{authblk}
\usepackage{blindtext}
\usepackage{titletoc}
\usepackage{afterpage}
\usepackage{authblk}
\usepackage{capt-of}
\usepackage{float}

\definecolor{DarkGreen}{rgb}{0,0.40,0}
\definecolor{FireBrick}{rgb}{0.698,0.133,0.133}
\definecolor{purple}{rgb}{0.5,0,0.5}

\usepackage[nameinlink,capitalize,noabbrev]{cleveref}

\usepackage{array}
\usepackage{makecell}

\def\half{{\frac{1}{2}}}

\newcommand{\bea}{\begin{eqnarray}}
\newcommand{\eea}{\end{eqnarray}}

\usepackage{slashed}

\def\({\left(}
\def\){\right)}
\def\[{\left[}
\def\]{\right]}

\usepackage{dashbox}
\usepackage{xcolor}
\usepackage{colortbl}
\usepackage{arydshln}
\definecolor{lightyellow}{rgb}{1.0, 0.95, 0.7}
\definecolor{blue}{rgb}{0.0, 0.4, 1.0}
\definecolor{Blue}{rgb}{0,0,1}
\definecolor{darkgreen}{rgb}{0,0.40,0}
\definecolor{firebrick}{rgb}{0.698,0.133,0.133}
\definecolor{DarkGreen}{rgb}{0,0.40,0}
\definecolor{FireBrick}{rgb}{0.698,0.133,0.133}

\definecolor{colorA}{rgb}{1,0,0}
\definecolor{colorB}{rgb}{0,0.3,1}
\definecolor{colorC}{rgb}{0.9,0.8,0.2}
\definecolor{colorD}{rgb}{0,0.65,0}
\definecolor{lesslightgray}{rgb}{0.5,0.5,0.5}
\definecolor{light-gray}{gray}{0.95}

\let\hat\widehat

\newcommand{\calO}{\mathcal{O}}

\newcommand{\calT}{\mathcal{T}}

\newcommand{\bA}{\mathbf{A}}
\newcommand{\bB}{\mathbf{B}}
\newcommand{\bC}{\mathbf{C}}

\newcommand{\bE}{\mathbf{E}}

\newcommand{\bQ}{\mathbf{Q}}
\newcommand{\bR}{\mathbf{R}}

\newcommand{\bW}{\mathbf{W}}
\newcommand{\bX}{\mathbf{X}}
\newcommand{\bY}{\mathbf{Y}}

\newcommand{\ba}{\mathbf{a}}
\newcommand{\bb}{\mathbf{b}}
\newcommand{\bc}{\mathbf{c}}

\newcommand{\bp}{\mathbf{p}}

\newcommand{\bx}{\mathbf{x}}

\newcommand{\bz}{\mathbf{z}}
\newcommand{\bxi}{{\bm{\xi}}}

\newcommand{\Max}{\mathop{\rm Max}}
\newcommand{\Min}{\mathop{\rm Min}}
  
\newcommand{\argmax}{\mathop{\mathrm{ArgMax}}}

\newcommand{\Softmax}{\mathop{\rm{Softmax}}}
\newcommand{\Sparsemax}{\mathop{\rm{Sparsemax}}}

\newcommand{\lse}{\mathop{\rm{lse}}}
\newcommand{\alphaentmax}{\mathop{\alpha\text{-}\mathrm{EntMax}}}

\newcommand{\sT}{ \mathsf{T} }

\newcommand{\sumM}{\sum_{\mu=1}^M}

\def\R{\mathbb{R}}

\let\cite\citep 

\usepackage{amsthm}
\makeatletter
\def\th@remark{%
  \thm@headfont{\bfseries}%
  \normalfont %
  \thm@preskip\topsep \divide\thm@preskip\tw@
  \thm@postskip\thm@preskip
}
\makeatother
\usepackage[many]{tcolorbox}
\theoremstyle{definition}
\newtheorem{theorem}{Theorem}[section]
\tcolorboxenvironment{theorem}{
  breakable,
  colback=black!10,
  colframe=white,%
  width=\dimexpr\linewidth+10pt\relax,%
  enlarge left by=-5pt,%
  enlarge right by=-5pt,%
  boxsep=5pt,%
  boxrule=0pt,
  left=0pt,right=0pt,top=0pt,bottom=0pt,
  sharp corners,
  before skip=\topsep,
  after skip=\topsep
}
\newtheorem{lemma}{Lemma}[section]
\tcolorboxenvironment{lemma}{
  breakable,
  colback=black!10,
  colframe=white,%
  width=\dimexpr\linewidth+10pt\relax,%
  enlarge left by=-5pt,%
  enlarge right by=-5pt,%
  boxsep=5pt,%
  boxrule=0pt,
  left=0pt,right=0pt,top=0pt,bottom=0pt,
  sharp corners,
  before skip=\topsep,
  after skip=\topsep
}
\newtheorem{corollary}{Corollary}[theorem]
\tcolorboxenvironment{corollary}{
  breakable,
  colback=black!10,
  colframe=white,%
  width=\dimexpr\linewidth+10pt\relax,%
  enlarge left by=-5pt,%
  enlarge right by=-5pt,%
  boxsep=5pt,%
  boxrule=0pt,
  left=0pt,right=0pt,top=0pt,bottom=0pt,
  sharp corners,
  before skip=\topsep,
  after skip=\topsep
}

\theoremstyle{definition}
\newtheorem{definition}{Definition}[section]
\tcolorboxenvironment{definition}{
  breakable,
  colback=black!10,
  colframe=white,%
  width=\dimexpr\linewidth+10pt\relax,%
  enlarge left by=-5pt,%
  enlarge right by=-5pt,%
  boxsep=5pt,%
  boxrule=0pt,
  left=0pt,right=0pt,top=0pt,bottom=0pt,
  sharp corners,
  before skip=\topsep,
  after skip=\topsep
}
\theoremstyle{remark}
\newtheorem{remark}{Remark}[section]

\tcolorboxenvironment{assumption}{
  breakable,
  colback=black!10,
  colframe=white,%
  width=\dimexpr\linewidth+10pt\relax,%
  enlarge left by=-5pt,%
  enlarge right by=-5pt,%
  boxsep=5pt,%
  boxrule=0pt,
  left=0pt,right=0pt,top=0pt,bottom=0pt,
  sharp corners,
  before skip=\topsep,
  after skip=\topsep
}

\tcolorboxenvironment{problem}{
  breakable,
  colback=black!10,
  colframe=white,%
  width=\dimexpr\linewidth+10pt\relax,%
  enlarge left by=-5pt,%
  enlarge right by=-5pt,%
  boxsep=5pt,%
  boxrule=0pt,
  left=0pt,right=0pt,top=0pt,bottom=0pt,
  sharp corners,
  before skip=\topsep,
  after skip=\topsep
}

\crefname{theorem}{Theorem}{Theorems}
\crefname{proposition}{Proposition}{Propositions}
\crefname{lemma}{Lemma}{Lemmas}
\crefname{corollary}{Corollary}{Corollaries}
\crefname{definition}{Definition}{Definitions}
\crefname{assumption}{Assumption}{Assumptions}
\crefname{remark}{Remark}{Remarks}
\crefname{problem}{Problem}{Problems}
\crefname{property}{Property}{property}

\numberwithin{equation}{section}
\numberwithin{theorem}{section}
\numberwithin{proposition}{section}
\numberwithin{definition}{section}
\numberwithin{lemma}{section}
\numberwithin{assumption}{section}
\numberwithin{remark}{section}

\usepackage{lipsum}

\makeatletter
\let\save@mathaccent\mathaccent
\newcommand*\if@single[3]{%
    \setbox0\hbox{${\mathaccent"0362{#1}}^H$}%
    \setbox2\hbox{${\mathaccent"0362{\kern0pt#1}}^H$}%
    \ifdim\ht0=\ht2 #3\else #2\fi
}
\newcommand*\rel@kern[1]{\kern#1\dimexpr\macc@kerna}
\newcommand*\widebar[1]{\@ifnextchar^{{\wide@bar{#1}{0}}}{\wide@bar{#1}{1}}}
\newcommand*\wide@bar[2]{\if@single{#1}{\wide@bar@{#1}{#2}{1}}{\wide@bar@{#1}{#2}{2}}}
\newcommand*\wide@bar@[3]{%
    \begingroup
    \def\mathaccent##1##2{%
        \let\mathaccent\save@mathaccent
        \if#32 \let\macc@nucleus\first@char \fi
        \setbox\z@\hbox{$\macc@style{\macc@nucleus}_{}$}%
        \setbox\tw@\hbox{$\macc@style{\macc@nucleus}{}_{}$}%
        \dimen@\wd\tw@
        \advance\dimen@-\wd\z@
        \divide\dimen@ 3
        \@tempdima\wd\tw@
        \advance\@tempdima-\scriptspace
        \divide\@tempdima 10
        \advance\dimen@-\@tempdima
        \ifdim\dimen@>\z@ \dimen@0pt\fi
        \rel@kern{0.6}\kern-\dimen@
        \if#31
        \overline{\rel@kern{-0.6}\kern\dimen@\macc@nucleus\rel@kern{0.4}\kern\dimen@}%
        \advance\dimen@0.4\dimexpr\macc@kerna
        \let\final@kern#2%
        \ifdim\dimen@<\z@ \let\final@kern1\fi
        \if\final@kern1 \kern-\dimen@\fi
        \else
        \overline{\rel@kern{-0.6}\kern\dimen@#1}%
        \fi
    }%
    \macc@depth\@ne
    \let\math@bgroup\@empty \let\math@egroup\macc@set@skewchar
    \mathsurround\z@ \frozen@everymath{\mathgroup\macc@group\relax}%
    \macc@set@skewchar\relax
    \let\mathaccentV\macc@nested@a
    \if#31
    \macc@nested@a\relax111{#1}%
    \else
    \def\gobble@till@marker##1\endmarker{}%
    \futurelet\first@char\gobble@till@marker#1\endmarker
    \ifcat\noexpand\first@char A\else
    \def\first@char{}%
    \fi
    \macc@nested@a\relax111{\first@char}%
    \fi
    \endgroup
    }
\makeatother
\let\bar\widebar

\makeatletter
\newcommand*{\redefinesymbolwitharg}[1]{%
  \expandafter\let\csname ltx#1\expandafter\endcsname\csname #1\endcsname
  \@namedef{#1}{\@ifnextchar{^}{\@nameuse{#1@}}{\@nameuse{#1@}^{}}}%
  \expandafter\def\csname #1@\endcsname^##1##2{%
     \csname ltx#1\endcsname\ifx!##1!\else^{##1}\fi\mathopen{}\mathclose\bgroup\left(##2\aftergroup\egroup\right)
     }%
}
\makeatother
\redefinesymbolwitharg{sin}
\redefinesymbolwitharg{cos}

\definecolor{LightCyan}{rgb}{0.8, 0.9, 1}
\definecolor{LightGray}{rgb}{0.83, 0.83, 0.83}
\newcolumntype{b}{>{\columncolor{LightCyan}\hspace{0pt}}c}

\newcolumntype{g}{>{\columncolor{LightGray}\hspace{0pt}}c}

\usepackage{url}
\usepackage{amsmath}
\usepackage{amssymb}
\usepackage{mathtools}
\usepackage{amsthm}

\usepackage[accepted]{icml2024}

\usepackage[textsize=tiny]{todonotes}

\icmltitlerunning{\textsc{BiSHop}: Bi-Directional Cellular Learning for Tabular Data with Generalized Sparse Modern Hopfield Model}

\setlist[itemize]{leftmargin=1em}
\setlist[enumerate]{leftmargin=1.4em}

\begin{document}

\twocolumn[
\icmltitle{\textsc{BiSHop}: Bi-Directional Cellular Learning for Tabular Data\\ with Generalized Sparse Modern Hopfield Model}

\icmlsetsymbol{equal}{*}

\begin{icmlauthorlist}
\icmlauthor{Chenwei Xu}{equal,nucs}
\icmlauthor{Yu-Chao Huang}{equal,twupt}
\icmlauthor{Jerry Yao-Chieh Hu}{equal,nucs}
\icmlauthor{Weijian Li}{nucs}
\icmlauthor{Ammar Gilani}{nucs}
\icmlauthor{Hsi-Sheng Goan}{twupt,twuquan,twupd}
\icmlauthor{Han Liu}{nucs,nustat}
\end{icmlauthorlist}

\icmlaffiliation{nucs}{Department of Computer Science, Northwestern University, Evanston, IL, USA}
\icmlaffiliation{nustat}{Department of Statistics and Data Science, Northwestern University, Evanston, IL, USA}
\icmlaffiliation{twuquan}{Center for Quantum Science and Engineering, National Taiwan University, Taipei, Taiwan}
\icmlaffiliation{twupd}{Physics Division, National Center for Theoretical Sciences, Taipei, Taiwan
}
\icmlaffiliation{twupt}{Department of Physics and Center for Theoretical Physics, National Taiwan University, Taipei, Taiwan}

\icmlcorrespondingauthor{Chenwei Xu}{\href{mailto:cxu@northwestern.edu}{cxu@northwestern.edu}}
\icmlcorrespondingauthor{Yu-Chao Huang}{\href{mailto:r11222015@ntu.edu.tw}{r11222015@ntu.edu.tw}}
\icmlcorrespondingauthor{Jerry Yao-Chieh Hu}{\href{mailto:jhu@u.northwestern.edu}{jhu@u.northwestern.edu}}
\icmlcorrespondingauthor{Weijian Li}{\href{mailto: weijianli@u.northwestern.edu}{weijianli@u.northwestern.edu}}
\icmlcorrespondingauthor{Ammar Gilani}{\href{mailto:ammargilani2024@u.northwestern.edu}{ammargilani2024@u.northwestern.edu}}
\icmlcorrespondingauthor{Hsi-Sheng Goan}{\href{mailto:goan@phys.ntu.edu.tw}{goan@phys.ntu.edu.tw}}
\icmlcorrespondingauthor{Han Liu}{\href{mailto:hanliu@northwestern.edu}{hanliu@northwestern.edu}}

\icmlkeywords{Machine Learning, ICML}

\vskip 0.3in
]

\printAffiliationsAndNotice{\icmlEqualContribution} %
\titlespacing*{\section}{0pt}{0pt}{0pt}
\titlespacing*{\subsection}{0pt}{0pt}{0pt}
\titlespacing*{\subsubsection}{0pt}{0pt}{0pt}

\begin{abstract}
We introduce the \textbf{B}i-Directional \textbf{S}parse \textbf{Hop}field Network (\textbf{BiSHop}), a novel end-to-end framework for tabular learning. 
BiSHop handles the two major challenges of deep tabular learning: 
non-rotationally invariant data structure and feature sparsity in tabular data.
Our key motivation comes from the recently established connection between associative memory and attention mechanisms. 
Consequently, BiSHop uses a dual-component approach, sequentially processing data both column-wise and row-wise through two interconnected directional learning modules. 
Computationally,  these modules house layers of generalized sparse modern Hopfield layers, a sparse extension of the modern Hopfield model with learnable sparsity.  
Methodologically,  BiSHop facilitates multi-scale representation learning, capturing both intra-feature and inter-feature interactions, with adaptive sparsity at each scale.
Empirically,  through experiments on diverse real-world datasets, BiSHop surpasses current SOTA methods with significantly fewer HPO runs, marking it a robust solution for deep tabular learning.
The code is available on \href{https://github.com/MAGICS-LAB/BiSHop}{GitHub}; future updates are on \href{https://arxiv.org/abs/2404.03830}{arXiv}.

\end{abstract}

\section{Introduction}
\label{sec:intro}
The field of developing deep learning architectures for tabular data has experienced rapid advancements \cite{somepalli2021saint,gorishniy2021revisiting,arik2021tabnet,huang2020tabtransformer}. 
The primary driving force behind this trend is to overcome the limitations of the current dominant methods for tabular data: tree-based methods.
Although tree-based methods excel in tabular learning, they cannot integrate with deep learning architectures.
Therefore, pursuing deep tabular learning is not just a matter of enhancing performance but is also crucial to bridging the existing gap.
Recent research has proposed new methodologies in deep tabular learning. However, a recent tabular benchmark study \cite{grinsztajn2022tree} reveals that tree-based methods still surpass deep learning models. This is due to two main challenges in deep tabular learning, as highlighted by \citet[Section 5.3 \& 5.4]{grinsztajn2022tree}:
\begin{enumerate}[leftmargin=2.3em]
\item  [(C1)] 
    \label{item:C1}
    \textbf{Non-Rotationally Invariant Data Structure:} The non-rotationally invariant structure of tabular data weakens the effectiveness of deep learning models that have rotationally invariant learning procedures.
    \item  [(C2)] 
    \label{item:C2}
    \textbf{Feature Sparsity:} Tabular datasets are generally sparser than typical datasets used in deep learning, which makes it challenging for deep learning models to learn from uninformative features.
\end{enumerate}

Inspired by the hierarchical and interconnected nature of the human brain \cite{presigny2022colloquium,krotov2021hierarchical}, 
we introduce the \textbf{Bi}-Directional \textbf{S}parse \textbf{Hop}field Network (\textbf{BiSHop}), a Hopfield-based deep learning framework tailored for tabular data.
To address the non-rotationally invariant data structure of tabular data \hyperref[item:C1]{(C1)}, our model employs a dual-component design, named the Bi-directional Sparse Hopfield Module (BiSHopModule). 
This design mirrors the human brain's memory mechanisms, where different regions work collaboratively to form and retrieve associative memories. 
Our model uses bi-directional learning through two separate Hopfield models, focusing on column-wise and row-wise patterns separately. This approach incorporates the tabular data's inherent structure as an inductive bias.

For tackling the features sparsity in tabular data \hyperref[item:C2]{(C2)}, 
we utilize the generalized sparse modern Hopfield model \cite{wu2023stanhop}. 
The generalized sparse modern Hopfield model is an extension to the sparse modern Hopfiled model \cite{hu2023sparse} and modern Hopfiled model \cite{ramsauer2020hopfield} with the learnable sparsity.
It offers robust representation learning and seamlessly integrates with existing deep learning architectures, ensuring focus on crucial information. 
Furthermore, inspired by brain's multi-level organization of associative memory, we stack multiple layers of the generalized sparse modern Hopfield model within BiSHopModule.
As a result, each layer learns representations at unique scales, adjusting its sparsity accordingly, adding \hyperref[item:C2]{(C2)} as another inductive bias to the model.

At its core, BiSHop facilitates multi-scale representation learning \cite{scl+22}, capturing both intra-feature and inter-feature dynamics while adjusting sparsity for each scale. 
The model identifies representations across various scales in all directions, whether column-wise or row-wise.
These refined representations are subsequently concatenated for downstream inference, ensuring a holistic bi-directional learning approach tailored for tabular data.

\begin{figure*}
    \centering
    \includegraphics[scale=0.75]{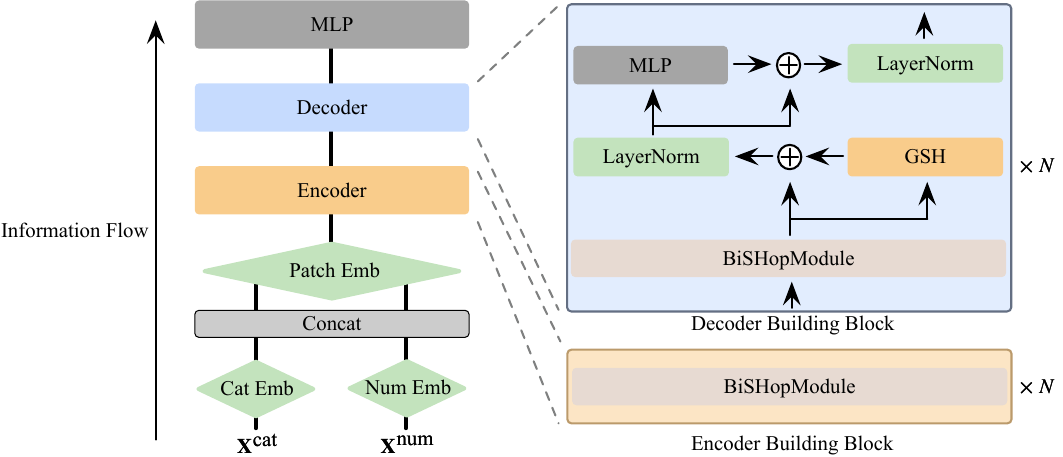}
    \caption{\small
    High-Level Visualization of BiSHop's Pipeline.
    }
    \label{fig:energy}
\end{figure*}

\paragraph{Contributions.}
Our contributions are twofold:
\begin{itemize}
    \item 
    Methodologically, we propose BiSHop, a novel deep-learning model for tabular data.
    BiSHop integrates two inductive biases (\hyperref[item:C1]{C1}, \hyperref[item:C2]{C2}) using the BiSHopModule and a hierarchical learning structure.
    The BiSHopModule utilizes the generalized sparse modern Hopfield model \cite{wu2023stanhop} for tabular feature learning, enabling multi-scale sparsity learning with superior noise-robustness. 
    We also present a hierarchical two-joint design to handle the intrinsic structure of tabular data with learnable sparsity and multi-scale cellular learning. 
    Additionally, we adopt tabular embedding \cite{huang2020tabtransformer, gorishniy2021revisiting, gorishniy2022embeddings} to enhance representation learning for both numerical and categorical features. 
    
    \item Experimentally, 
    we conduct comprehensive experiments on diverse real-world datasets as well as a tabular benchmark \cite{grinsztajn2022tree}. 
    This encompasses a total of 18 classification tasks and 11 regression tasks. 
    We compare BiSHop with both SOTA tree-based and deep learning methods.
    Our results show that BiSHop outperforms baselines across most of tested datasets, including both regression and classification tasks.
\end{itemize}

\paragraph{Notations.}
We denote vectors by lowercase bold letters, and
matrices by upper bold letters
For vectors 
$\ba$, $\bb$, we define their inner product as $\Braket{\ba,\bb} = \ba^\sT \bb$.
We use the shorthand $[I]$ to represent the index set $\{1,\cdots,I\}$ with $I$ being a positive integer. 
For matrices, we denote the spectral norm as $\norm{\cdot}$, which aligns with the $l_2$-norm for vectors. 
We denote the memory patterns by $\bxi\in\R^d$ and the query pattern by $\bx\in\R^d$, and $\bm{\Xi}\coloneqq \[\bxi_1,\cdots,\bxi_M\]\in \R^{d\times M}$ 
as shorthand for memory patterns $\{\bxi_\mu\}_{\mu\in[M]}$.

\section{Background: Dense and Generalized Sparse Modern Hopfield Model}
\label{sec:model}
This section provides a concise overview of the modern Hopfield model \cite{ramsauer2020hopfield} and the generalized sparse modern Hopfield model \cite{wu2023stanhop}.
\citet{wu2023stanhop} presents an extension to \cite{hu2023sparse,ramsauer2020hopfield}, utilizing the Tsallis $\alpha$-entropy \cite{tsallis1988possible}

\subsection{(Dense) Modern Hopfield Models}

Let $\bx \in \mathbb{R}^d$ be the query pattern and $\bm{\Xi} = [\bxi_1, \cdots, \bxi_M] \in \mathbb{R}^{d \times M}$ the memory patterns. 
The aim of Hopfield models \cite{hopfield1982neural,hopfield1984neurons,krotov2016dense,demircigil2017model,krotov2020large} is to store these memory patterns $\bm{\Xi}$ and retrieve a specific memory $\bxi_\mu$ when given a query $\bx$. 
These models comprise two primary components: an \textit{energy function} $E(\bx)$ that encodes memories into its local minima, and a \textit{retrieval dynamics} $\calT(\bx)$ that fetches a memory by iteratively minimizing $E(\bx)$ starting with a query.

\citet{ramsauer2020hopfield} propose the (dense/vanilla) modern Hopfield model with a specific set of $E$ and $\calT$, and integrate it into deep learning architectures via its  connection with attention mechanism, offering enhanced performance, and theoretically guaranteed exponential memory capacity.
Specifically, they introduce a Hopfield energy function:
\begin{align}
\label{eqn:MHM}
    E(\bx) = -\lse(\beta,\bm{\Xi}^\sT \bx) + \frac{1}{2} \Braket{\bx,\bx} ,
\end{align}
and the corresponding memory retrieval dynamics:  
\begin{align*}
\calT_{\text{Dense}}(\bx) = \bm{\Xi} \cdot \Softmax(\beta \bm{\Xi}^\sT \bx) = \bx^{\text{new}}.
\end{align*}
The function $\lse\(\beta,\bz\)\coloneqq \log\(\sumM \exp{\beta z_\mu}\)/\beta$ is the log-sum-exponential for any given vector $\bz\in\R^M$ and $\beta>0$.
Surprisingly, their findings reveal:
\begin{itemize}
    \item The $\calT_{\text{Dense}}$ dynamics converge to memories provably and retrieve patterns accurately in just one step.
    \item The modern Hopfield model from \eqref{eqn:MHM} possesses an exponential memory capacity in pattern size $d$.
    \item Notably, the one-step approximation of  $\calT_{\text{Dense}}$ mirrors the attention mechanism in transformers, leading to a novel architecture design: the Hopfield layers.
\end{itemize}

\subsection{Generalized Sparse Modern Hopfield Model}

\label{sec:GSHM}
This section follows \citep[Section~3]{wu2023stanhop}.
For self-containedness, we also summarize the useful theoretical results of \cite{wu2023stanhop} in \cref{sec:sup_theory}.

\paragraph{Associative Memory Model.}
Let $\bz,\bp\in\R^M$, and $\Delta^{M}\coloneqq\{\bp\in\R^M_+ \mid \sum_\mu^M p_\mu=1\}$ be the $(M-1)$-dimensional unit simplex.
\citet{wu2023stanhop} introduce the generalized sparse Hopfield energy as a new associative memory model
\begin{align}
\label{eqn:GSH_energy}
E(\bx)=-\Psi^\star\(\beta \bm{\bm{\Xi}}^\sT \bx\) +\half \Braket{\bx,\bx},
\end{align}
where $\Psi^\star(\bz) \coloneqq \int\dd\bz \alphaentmax(\bz)$, and $\alphaentmax(\cdot)$ is defined as follows.
\begin{definition}[\cite{peters2019sparse}] 
\label{def:entmax}
The variational form of $\alphaentmax$ is defined as 
\begin{align}
   \label{eqn:GSH_retreival_dyn}
\alphaentmax(\bz) \coloneqq \argmax_{\bp \in \Delta^M} [\bp^\sT \bz-\Psi^{\alpha}(\bp)], 
\end{align}

where $\Psi^{\alpha}(\cdot)$ is the Tsallis entropic regularizer
\begin{align*}
    \Psi^{\alpha}(\bp)\coloneqq
\begin{cases}
    \frac{1}{\alpha(\alpha-1)}\sum^M_{\mu=1}\(p_\mu-p_\mu^\alpha\), \; & \alpha\neq 1,\\
    \-\sumM p_\mu\ln p_\mu,\quad &\alpha=1,
\end{cases}\;\;\text{for }\alpha\ge 1.\nonumber
\end{align*}

\end{definition}
The corresponding memory retrieval dynamics is given as
\begin{lemma}[Retrieval Dynamics, Lemma~3.2 of \cite{wu2023stanhop}]
\label{lemma:retrieval_dyn}
Given $t$ as the iteration number, the generalized sparse modern Hopfield model exhibits a retrieval dynamic
\begin{align}
\label{eqn:retrieval_dyn}
\calT(\bx_t) = \bm{\Xi}\alphaentmax( \beta \bm{\Xi}^\sT \bx_t ) = \bx_{t+1},
\end{align}
which ensures a monotonic decrease of the energy \eqref{eqn:GSH_energy}.
\end{lemma}
This model also enjoys nice memory retrieval properties:
\begin{lemma}[Convergence of Retrieval Dynamics $\calT$, Lemma~3.3 of \cite{wu2023stanhop}]
\label{lemma:convergence_sparse}
Given the energy function $E$ and retrieval dynamics $\calT$ defined in \eqref{eqn:GSH_energy} and \eqref{eqn:GSH_retreival_dyn}, respectively.
For any sequence $\{\mathbf{x}_t\}_{t=0}^{\infty}$ generated by the iteration $\mathbf{x}_{t'+1} = \mathcal{T}(\mathbf{x}_{t'})$, all limit points of this sequence are stationary points of $E$.
\end{lemma}
\cref{lemma:convergence_sparse} ensures the (asymptotically) exact memory retrieval of this model (\eqref{eqn:GSH_energy} and \eqref{eqn:retrieval_dyn}),
Thus, it serves as a well-defined associative memory model.

In essence, \citet{wu2023stanhop} present this sparse extension of the modern Hopfield model through a construction of both $E$ and $\calT$ by convex conjugating the Tsallis entropic regularizers. 
This model not only adheres to the conditions for a well-defined modern Hopfield model, but also equips greater robustness (\cref{coro:noise}) and retrieval speed (\cref{thm:eps_sparse_dense} and \cref{coro:faster_convergence}) than the modern Hopfield model \cite{ramsauer2020hopfield}, see \cref{sec:theoretical_properties} for details.
In \cref{fig:feature},
we also provide proof-of-concept experimental validations on tabular datasets for \cref{thm:eps_sparse_dense}, \cref{coro:faster_convergence} and \cref{coro:noise}.

\paragraph{Generalized Sparse Modern Hopfield (GSH) Layers for Deep Learning.} 
Importantly, the generalized sparse modern Hopfield model serves as a valuable component in deep learning due to its connection to the transformer attention akin to its cousins.
Next, we review such connections and the Generalized Sparse Modern Hopfield (GSH) layers.

Following \cite{wu2023stanhop,hu2023sparse,ramsauer2020hopfield}, $\mathbf{X}$ and $\bm{\Xi}$ are defined in the associative space, embedded from the raw query $\mathbf{R}$ and memory patterns $\mathbf{Y}$, respectively, using $\mathbf{X}^\top=\mathbf{R}\mathbf{W}_Q\coloneqq \mathbf{Q}$ and $\bm{\Xi}^\top=\mathbf{Y} \mathbf{W}_K \coloneqq \mathbf{K}$ with matrices $\mathbf{W}_Q$ and $\mathbf{W}_K$. 
By transposing $\mathcal{T}$ from \eqref{eqn:retrieval_dyn} and applying $\mathbf{W}_V$ such that $\mathbf{V}\coloneqq \mathbf{K}\mathbf{W}_V$, we obtain:
\begin{align}
    \mathbf{Z}\coloneqq \mathbf{Q}^{\text{new}} \mathbf{W}_V =\alphaentmax(\beta  \mathbf{Q}\mathbf{K}^\top)\mathbf{V},
\end{align}
introducing an attention mechanism with the $\alphaentmax$ activation function. Substituting $\mathbf{R}$ and $\mathbf{Y}$ back in, the Generalized Sparse Modern Hopfield ($\mathtt{GSH}$) layer is formulated as:
\begin{align*}
\mathtt{GSH}(\mathbf{R},\mathbf{Y})= \alphaentmax(\beta  {\mathbf{R}}\mathbf{W}_Q \mathbf{W}_K^\top\mathbf{Y}^\top)\mathbf{Y}\mathbf{W}_K \mathbf{W}_V.
\end{align*}
This allows the seamless integration of the generalized sparse modern Hopfield model into deep learning architectures.
Concretely, the $\mathtt{GSH}$ layer takes matrices $\bR$, $\bY$ as inputs, with the weight matrices $\bW_Q$, $\bW_K$, $\bW_V$. 
Depending on its configuration, it offers several functionalities:
\begin{enumerate}

    \item \textbf{Memory Retrieval:}
    In this learning-free setting, weight matrices $\bW_K$, $\bW_Q$, and $\bW_V$ are set as identity matrices. Here, $\bR$ represents the query input, and $\bY$ denotes the stored memory patterns for retrieval.

    \item \textbf{$\mathtt{GSH}$:}
    This configuration takes $\bR$ and $\bY$ as inputs. 
    Intending to substitute the attention mechanism, the weight matrices $\bW_K$, $\bW_Q$, and $\bW_V$ are rendered learnable. 
    Furthermore, $\bR$, $\bY$, and $\bY$ serve as the sources for query, key, and value respectively. To achieve a self-attention-like mechanism, $\bR$ is set equal to $\bY$.

    \item \textbf{$\mathtt{GSHPooling}$:}
    With inputs $\bQ$ and $\bY$, this layer uses $\bQ$ as a static \textbf{prototype pattern}, while $\bY$ contains patterns over which pooling is desired. 
    Given that the query pattern is replaced by the static prototype pattern $\bQ$, the only learnable weight matrices are $\bW_K$ and $\bW_V$.

    \item \textbf{$\mathtt{GSHLayer}$:}
    The $\mathtt{GSHLayer}$ takes the query $\bR$ as its single input. 
    It includes learnable weight matrices $\bW_K$ and $\bW_V$, which function as our stored patterns and their corresponding projections.
    This design ensures that our key and value are decoupled from the input. 
    In practice, we set $\bW_Q$ and $\bY$ as identity matrices.

\end{enumerate}
In this work, we utilize \textbf{$\mathtt{GSH}$} and \textbf{$\mathtt{GSHPooling}$} layers\footnote{\url{https://github.com/MAGICS-LAB/STanHop}}.

\section{Methodology}
\label{sec:method}

As in \cref{fig:energy}, BiSHop uses three distinct parts to integrate two pivotal inductive biases in tabular data: non-rotationally invariant data structures \hyperref[item:C1]{(C1)} and sparse information in features \hyperref[item:C2]{(C2)}  \citep[Section 5.3 \& 5.4]{grinsztajn2022tree}:
\begin{itemize}
    \item 
    A joint \textbf{Tabular Embedding} layer is designed to process categorical and numerical data separately.
    \item 
    The \textbf{Bi-Directional Sparse Hopfield Module (BiSHopModule)} leverages the generalized sparse modern Hopfield model.
    This module incorporates the non-rotationally invariant bias through two interconnected $\mathtt{GSH}$ blocks for row-wise and column-wise learning.
    \item 
    \textbf{Stacked BiSHopModules} for hierarchical learning, addressing sparse features. 
    Each layer in the stack module captures information at different scales, allowing for scale-specific sparsity.
\end{itemize}
We provide a detailed breakdown of each part as follows.

\begin{figure*}[h]
    \centering
    \includegraphics[scale=0.53]{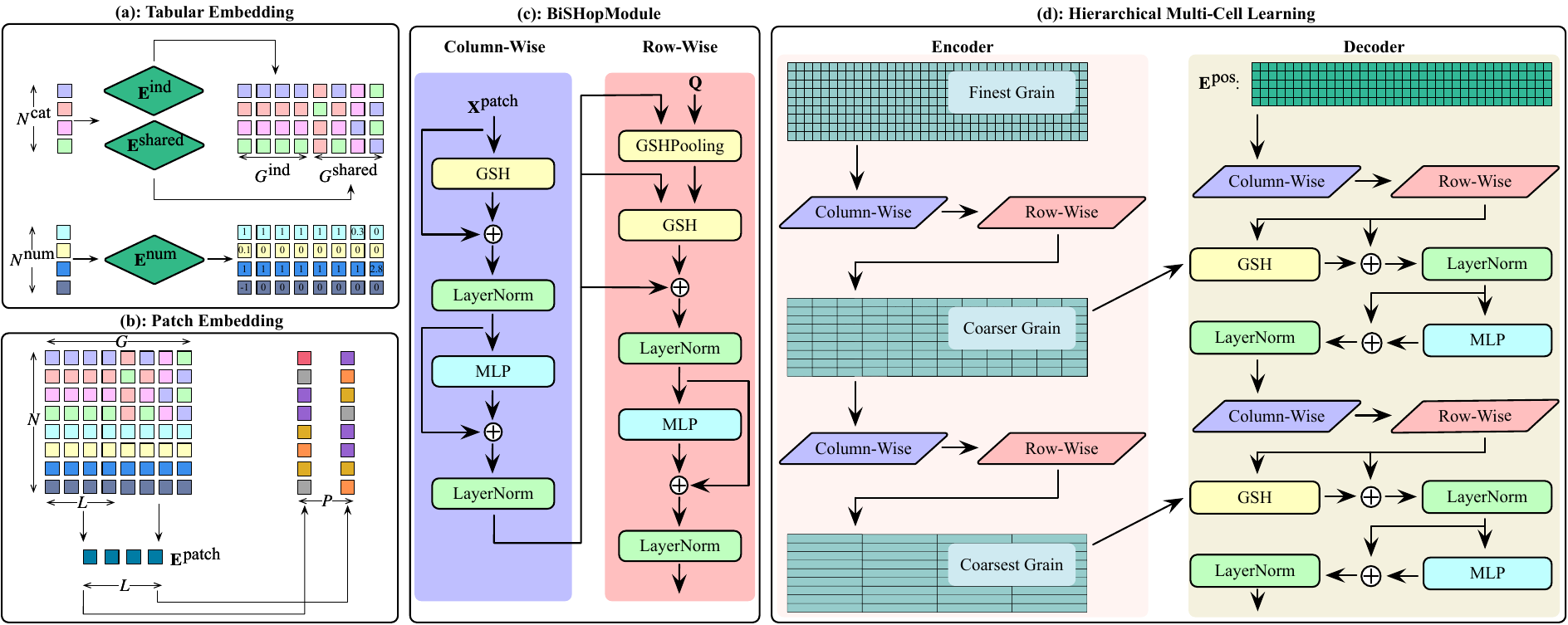}
    \caption{\small
\textbf{BiSHop.}
\textbf{(a) Tabular Embedding:}
For a given input feature $\bx=(\bx^{\text{cat}},\bx^{\text{num}}) \in\R^{N = N^{\text{cat}} + N^{\text{num}} }$, the tabular embedding produces embeddings denoted as $\bE^{\text{emb}}(\bx)\in\R^{N\times G}$.
\textbf{(b) Patch Embedding:}
Using the combined numerical and categorical embeddings $\bE^{\text{emb}}(\bx)\in\R^{N \times G}$, the patch embedding gathers embedding information, subsequently reducing dimensionality from $G$ to $P =\lceil G/L \rceil$ for all $N$ features using a stride length of $L$.
\textbf{(c) BiSHopModule:}
The Bi-Directional Sparse Hopfield Module (BiSHopModule) leverages the generalized sparse modern Hopfield model. It integrates the tabular structure's inductive bias (C1) by deploying interconnected row-wise and column-wise $\mathtt{GSH}$ layers.
\textbf{(d) Hierarchical Cellular Learning Module:}
Employing a stacked encoder-decoder structure, we facilitate hierarchical cellular learning where both the encoder and decoder consist of the BiSHopModule across $H$ layers. 
This arrangement enables BiSHop to derive refined representations from both directions across multiple scales. 
These representations are then concatenated for downstream inference, ensuring a holistic bi-directional cellular learning specially tailored for tabular data.
}

\label{fig:flowchart}

\end{figure*}

\subsection{Tabular Embedding}

Tabular embedding consists of three parts: \textbf{categorical embedding} $\bE^{\text{cat}}$, \textbf{numerical embedding} $\bE^{\text{num}}$, and \textbf{patch embedding} $\bE^{\text{patch}}$. 
The categorical embedding not only learns the representations within individual categorical features but also capture the inter-relation among all categorical features. 
The numerical embedding represents each numerical feature with a one-hot-like representation and thus benefits neural network learning numerical features.
The patch embedding captures localized feature information by aggregating across feature dimensions, at the same time reducing computation overhead.
Starting from this section, we denote $\bx \in \mathbb{R}^{N}$ 
any given tabular data point with $N$ features. 
We suppose each $\bx$ has $N^{\text{num}} $ numerical feature $\bx^{\text{num}} $ and $N^{\text{cat}}$ categorical feature $\bx^{\text{cat}}$, where $\bx=(\bx^{\text{num}}, \bx^{\text{cat}})$. 
The categorical embedding $\bE^{\text{cat}}$ and numerical embedding $\bE^{\text{num}}$ transforms $\bx^{\text{cat}}$ and $\bx^{\text{num}}$ to a embedding dimension $G$, separately. 
The patch embedding $\bE^{\text{patch}}$ then reduces $G$ to the patch embedding dimension $P$.

\paragraph{Categorical Embedding.}
\label{Categorical Embedding}
For categorical embedding  $\bE^{\text{cat}}$, we use learnable column embedding proposed by \citet{huang2020tabtransformer}. 
For a tabular data point $\bx = (\bx^{\text{num}}, \bx^{\text{cat}})$, the column embedding only acts on the categorical features $\bx^{\text{cat}}$, denoted as $\bE^\text{cat}(\bx^{\text{cat}})$.
It comprises a shared embedding $\bE^\text{shared}(\bx^{\text{cat}})$ for all categorical features and $N^\text{cat}$ individual embeddings for each categorical feature $\{x^\text{cat}_i\}_{i \in [N^\text{cat}]}$, where $[N^\text{cat}] = \{1, \cdots, N^\text{cat}\}$.
We denote the shared embedding dimension as $G^\text{shared}$ and the individual embedding dimension as $G^\text{ind}$, where $G=G^\text{shared}+G^\text{ind}$.
The shared embedding $\bE^\text{shared}(\bx^{\text{cat}}) \in \mathbb{R}^{N^\text{cat}\times G^\text{shared}}$ represents each categorical feature differently. 
The individual embedding $\bE^\text{ind} = \{\bE^\text{ind}_1, \cdots, \bE^\text{ind}_{N^\text{cat}}\}$ represents each category within one categorical feature differently. 
Each individual embedding $\bE^\text{ind}_{i} (\cdot) \in \mathbb{R}^{1 \times G^\text{ind}}$ is a scalar-to-vector map acting on each categorical feature $\{x^\text{cat}_i\}_{i\in[N^\text{cat}]}$. 
To obtain the final categorical embedding, we first concatenate all individual embeddings row-wise:
$\bE^\text{ind}(\bx^\text{cat}) := \mathtt{Concat}([\bE^\text{ind}_1(x^\text{cat}_1), \ldots, \bE^\text{ind}_{N^\text{cat}}(x^\text{cat}_{N^\text{cat}})], \mathtt{axis} = 0) \in \mathbb{R}^{N^\text{cat} \times G^\text{ind}}.$
Then, we concatenate the shared embedding with all individual embeddings column-wise:
$\bE^\text{cat}(\bx^\text{cat}) := \mathtt{Concat}([\bE^\text{shared}(\bx^\text{cat}), \bE^\text{ind}(\bx^\text{cat})], \mathtt{axis} = 1) \in \mathbb{R}^{N^\text{cat} \times G}.$
$\bE^\text{ind}$ represents the unique category in each feature and $\bE^\text{shared}$ represents the unique feature.
$\bE^\text{cat}$ enables our model to capture both the relationship between each feature and each category, with the flexibility to train shared and individual components separately.

\paragraph{Numerical Embedding.}
We employ the numerical embedding method as described in \citet{gorishniy2021revisiting, gorishniy2022embeddings}. 
The numerical embedding $\bE^{\text{num}}$ only acts on the numerical features $\bx^{\text{num}}$, as in $\bE^\text{num}(\bx^{\text{num}}) \in \mathbb{R}^{N^\text{num}\times G}$. 
Given a numerical feature $\{x^\text{num}_i\}_{i\in[N^\text{num}]}$, the embedding process begins by determining $G$ quantiles.
To start, we determine $G$ quantiles for each numerical feature. 
Quantiles represent each numerical data distribution by dividing it into equal parts.
For a numerical feature $\{x^\text{num}_j\}_{j \in N^\text{num}}$, we first sort all its values in the training data, $\bx^\text{num}_j$, in ascending order.
Then, we split the sorted data into $G$ equal parts, where each part contains an equal fraction of the total data points.
We define the boundaries of these parts as $b_{j,0}, \cdots, b_{j,G}$, where $b_{j,0}$ is the smallest value in $\bx^{\text{num}}_j$.
We express the embedding for a specific value $x_j$ as a $G$-dimensional vector, $\bE^\text{num}_j(x_j) = (e_{j,1}, \cdots, e_{j,G})\in\R^G$. 
We compute the value of each $e_{j, g}$, where $1 \leq g \leq G$ according to the following function:
\begin{align*}
e_{j,g} :=
\begin{cases}
0, & \text{if } x_j < b_{j, g-1} \text{ and } g > 1,\\
1, & \text{if } x_j \geq b_{j, g} \text{ and } g < G,\\
\frac{x_j - b_{j, g-1}}{b_{j,g} - b_{j, g-1}}, & \text{otherwise}.
\end{cases}
\end{align*}
For the final embedding, 
we have $\bE^{\text{num}}(\bx^{\text{num}})\in\R^{N^{\text{num}}\times G}$.
We denote this numerical embedding as piece-wise linear embedding. 
This technique normalizes the scale of numerical features and captures the quantile information for each data point within the numerical feature.
It enhances the representation of numerical feature in deep learning.
Concatenating $\bE^{\text{num}}(\bx^{\text{num}})$ with $\bE^{\text{cat}}(\bx^{\text{cat}})$  row-wise, we obtain:
$\bE^{\text{emb}}(\bx)\coloneqq
\mathtt{Concat}\left( \bE^{\text{num}}(\bx^{\text{num}}), \bE^{\text{cat}}(\bx^{\text{cat}}) ,\mathtt{axis}=0 \right)$,
where $\bE^{\text{emb}}(\bx) \in\R^{N\times G}$.
Namely, we call each point $\bE_{n, g}^{\text{emb}}(\bx)$ as a single cell.
The categorical and numerical embedding is in \cref{fig:flowchart} (a).

\paragraph{Patch Embedding.}
Motivated by~\cite{Yuqietal-2023-PatchTST,zhang2023crossformer, qrsxz22}, we adopt patch embedding (shown in \cref{fig:flowchart} (b)) to enhance the awareness of both local and non-local patterns, capturing intricate details often missed at the single-cell level.
Specifically, we divide embeddings into patches that aggregate multiple cells. 
To simplify the computation process, we transpose the numerical and categorical embedding dimensions. 
For convenience, we denote the previous embedding outcomes as $\bX^{\text{emb}} \coloneqq (\bE^\text{emb}(\bx))^\top \in \mathbb{R}^{G \times N}$. 
The patch embedding $\bE^\text{patch}$ reduces the embedding dimension $G$ by a stride factor $L$, leading to a new and smaller patched embedding dimension $P :=  \lceil{G/L}\rceil$,
where $\lceil \cdot \rceil$ is the ceiling function.
Furthermore, we introduce a new embedding dimension $D^\text{model}$ to represent each patch's hidden states.
The patched embedding is $\bE^\text{patch}(\bX^{\text{emb}}) \in \mathbb{R}^{P \times N \times D^\text{model}}$. 
For future computation, we flip the patch dimension and feature dimension, resulting final output of patch embedding $\bX^{\text{patch}} \coloneqq (\bE^\text{patch}(\bX^{\text{emb}}))^\top \in \mathbb{R}^{N \times P \times D^\text{model}}$.
This patch embedding method enhances our model's ability to interpret and integrate detailed local and broader contextual information from the data, crucial for in-depth analysis in deep learning scenarios.
For the $\bX^{\text{patch}}$, we denote it as having $N$ rows (features) and $P$ columns (embeddings).

\subsection{Bi-Directional Sparse Hopfield Module}
\label{sec:BiSHopModule}
By drawing parallels with the intricate interplay of different parts in the brain \cite{presigny2022colloquium}, we present the core design of the BiSHop framework, the Bi-Directional Sparse Hopfield Module (BiSHopModule),
as visualized in \cref{fig:flowchart} (c).
The BiSHopModule incorporates the generalized sparse modern Hopfield model and integrates the inductive bias of tabular structure (C1) through a unique structure of stacked row-wise and column-wise $\mathtt{GSH}$ blocks.
Specifically, the row-wise $\mathtt{GSH}$ focuses on capturing the embedding details for individual features, whereas the column-wise $\mathtt{GSH}$ aggregates information across all features.
We denote $\bX^\text{patch}_{n,p}$, where $n \in [N]$ and $p \in [P]$, as the element in the $n$-th row (feature) and $p$-th column (embedding).

\paragraph{Column-Wise Block.}
The column-wise $\mathtt{GSH}$ block (purple block on the LHS of \cref{fig:flowchart} (c)) captures hidden information across the embedding dimension $P$ for each feature.
The process begins by passing the patch embeddings of the $n$-th row of $\bX^{\text{patch}}$, $\bX^{\text{patch}}_{n, :}$, where $n \in [N]$, to the $\mathtt{GSH}$ layer for self-attention. 
This process is then followed by adding the original patch embeddings (similar to the residual connection in the standard transformer).
Next, we pass the output above through one $\mathtt{LayerNorm}$ layer, a Multi-Layer Perception (MLP) layer, and another $\mathtt{LayerNorm}$ to obtain the final output of the column-wise block, $\bX^{\text{col}}$:
\begin{align*}
\hat{\bX}^{\text{patch}}_{n, :} &:= \mathtt{LayerNorm} \left( \bX^{\text{patch}}_{n, :} + \mathtt{GSH}(\bX^{\text{patch}}_{n, :}, \bX^{\text{patch}}_{n, :}) \right),\\
\bX^{\text{col}} &:= \mathtt{LayerNorm} \left( \hat{\bX}^{\text{patch}} + {\mathtt{MLP}}(\hat{\bX}^{\text{patch}}) \right),
\end{align*}
This sequence of operations ensures the effective transformation of the embeddings, facilitating the extraction of meaningful information from the feature space \cite{swl21, swl24}. 

\paragraph{Row-Wise Block.}
The row-wise $\mathtt{GSH}$ block (pink block on the RHS of \ref{fig:flowchart} (c)) captures information across the feature dimension $N$.
For each feature, we apply both $\mathtt{GSHPooling}$ and $\mathtt{GSH}$ layers to its embedding dimensions. 
Specifically, we use $C$ learnable pooling vectors in each feature dimension to aggregate information across all embedding dimensions, forming a pooling matrix $\bQ \in \mathbb{R}^{C \times P \times D^{\text{model}}}$.
We represent the pooling at $p$-th embedded dimension as the $p$-the columns of $\bQ$, $\bQ_{:,p}$, where $p\in[P]$. 
The process begins by pooling the row-wise output $\bX^\text{row}_{:,p}$ using $\bQ_{:,p}$ in the $\mathtt{GSHPooling}$ step. 
Next, we combine this pooled output with the row-wise output again, and add the row-wise output to the result. 
Following this, we pass the output through a $\mathtt{LayerNorm}$ layer, an MLP layer, and another $\mathtt{LayerNorm}$ layer.
This sequence of operations yields the final output of the row-wise block:
\begin{align*}
\hat{\bQ}_{:,p} &:= \mathtt{GSHPooling}(\bQ_{:,p}, \bX^{\text{col}}_{:,p}), \\
\hat{\bX}^{\text{row}}_{:,p}& := \mathtt{GSH}(\bX^{\text{col}}_{:,p}, \hat{\bQ}_{:,p}), \\
\bar{\bX}^{\text{row}} & := \mathtt{LayerNorm} (\hat{\bX}^{\text{row}} + \bX^{\text{col}}), \\
\bX^{\text{row}} & := \mathtt{LayerNorm} (\bar{\bX}^{\text{row}} + \mathtt{MLP}(\bar{\bX}^{\text{row}})) ,
\end{align*}
This $\bQ$ pooling matrix design aggregates information from all patch embedding dimensions, and by setting $C \ll N$, it significantly reduces computational complexity. 

Together with the row-wise block, we summarize the entire BiSHopModule as a function:
\begin{align*}
    \mathtt{BiSHopModule}(\cdot) \colon \mathbb{R}^{P \times N} \rightarrow \mathbb{R}^{P \times N},
\end{align*}
where the input is $\bX^\text{patch}$ and the output is $\bX^\text{row}$.

\subsection{Stacked BiSHopModules for Multi-Scale Learning with Scale-Specific Sparsity} 
Motivated by the human brain's multi-level organization of associative memory \cite{presigny2022colloquium,krotov2021hierarchical}, we utilize a hierarchical structure to learn multi-scale information similar to \cite{zhang2023crossformer,zhou2021informer}.
This is illustrated in \cref{fig:flowchart} (d). 
This structure consists of two main components: the encoder and the decoder, both of which incorporate $H$ layers of BiSHopModules. 
Specifically, the encoder captures coarser-grained information across different scales, while the decoder makes forecasts based on the information encoded by the encoder.

\paragraph{Encoder.} 
The encoder (pink block on LHS of \cref{fig:flowchart} (d)) encodes data at multiple levels of granularity.
To accomplish this multi-level encoding, we use $H$ stacked BiSHopModules. 
These modules help in processing and understanding the data from different perspectives.
We also employ a learnable merging matrix \cite{liu2021swin} to aggregate $r$ adjacent patches of $\bX^{\text{patch}}$.
We denote the merging matrix at layer $h \in [H]$ as $\bE^{\text{merge}}_h \in \mathbb{R}^{r \times 1}$, which refines its input embeddings to be coarser at each level.
We refer to $h$-th level encoder output as $\bX^{\text{enc}, h}$ and input as $\bX^{\text{enc}, h-1}$. 
Concretely, at level $h$, we use $\bE^{\text{merge}}_h$ to aggregate $r$ adjacent embedding vectors from $\bX^{\text{enc}, h-1}$, producing a coarser embedding $\hat{\bX}^{\text{enc}, h-1}$.
We then pass $\hat{\bX}^{\text{enc},h-1}$ through the BiSHopModules, resulting in the output encoded embedding, denoted as $\bX^{\text{enc}, h}$. 
It is worth noting that $\bX^{\text{enc}, 0} = \bX^{\text{patch}}$. 
This granularity-decreasing process is iteratively applied across all layers in $1\le h\le H$.
We summarize the merging procedure at level $h$ as:
\begin{align*}
&\hat{\bX}^{\text{enc}, h}_{n, p} := \bE^{\text{merge}}_h \left( \bX^{\text{enc}, h}_{n, r \times p}, \ldots,  \bX^{\text{enc}, h}_{n, r \times (p+1)} \right),   0 \leq p \leq \frac{P}{r^h}, \nonumber
\end{align*}
for $0\leq h\leq H-1$, and then
\begin{align*}
\bX^{\text{enc}, h} := \mathtt{BiSHopModule}(\hat{\bX}^{\text{enc}, h-1}), \quad\text{for }1\le h \le H.
\end{align*}
\paragraph{Decoder.}
The decoder (yellow block on RHS of \cref{fig:flowchart} (d)) captures information from each level of encoded data. 
To accomplish this, we utilize $H$ stacked BiSHopModules and employ a positional embedding matrix $\bE^{\text{pos}} \in \mathbb{R}^{P \times S}$ to extract encoded information for prediction, where $S$ represents the number of extracted feature used for future forecast.  
Specifically, at the first level, we use the learnable matrix $\bE^{\text{pos}}$ to decode $S$ different representations through BiSHopModules, obtaining $\bX^{\text{pos}, 0}$. 
We then pass $\bX^{\text{pos}, 0}$ through $\mathtt{GSH}$ with the corresponding encoded data, followed by the addition to the encoded data at the $h$-th level $\bX^{\text{enc}, h}$. 
Next, we process the output through one $\mathtt{LayerNorm}$ layer, one $\mathtt{MLP}$ layer, and another $\mathtt{LayerNorm}$  layer, as follows:
\begin{align*}
\bX^{\text{pos}, h} &:= 
\begin{cases}
\mathtt{BiSHopModule}(\bE^{\text{pos}}), & h = 0, \\
\mathtt{BiSHopModule}(\bX^{\text{dec}, h - 1}), & 1 \leq h \leq H. 
\end{cases}
\\
\hat{\bX}^{\text{dec}, h} &:= \mathtt{GSH}(\bX^{\text{pos}, h}, \bX^{\text{enc, h}}),  1 \leq h \leq H, \\
\bar{\bX}^{\text{dec} ,h} &:= \mathtt{LayerNorm} (\hat{\bX}^{\text{dec} ,h} + \bX^{\text{pos}, h}), \\
\bX^{\text{dec}, h} &:= \mathtt{LayerNorm} (\bar{\bX}^{\text{dec} ,h} + \mathtt{MLP}(\bar{\bX}^{\text{dec} ,h})).
\end{align*}
For the final prediction, we flatten $\bX^{\text{dec}, H}$ and pass it to a new MLP predictor.

\paragraph{Learnable Sparsity at Each Scale.}
Drawing inspiration from the dynamic sparsity observed in the human brain \cite{stokes2013dynamic,leutgeb2005progressive,willshaw1969non},
the parameter $\alpha$ for each $\mathtt{GSH}$ layer is a learnable parameter by design \cite{wu2023stanhop,correia2019adaptively}, which allows BiSHopModule to adapt to different sparsity for different resolutions.
Namely, the learned representations at each scale are equipped with scale-specific sparsity.

\section{Experimental Studies}
\label{sec:exp}

In this section, we compare BiSHop with SOTA tabular learning methods, following the tabular learning benchmark paper \cite{grinsztajn2022tree}.
We summarize our experimental results in  \cref{tab:main_results} and \cref{tab:main_results_pipeline}.

\subsection{Experimental Setting}
\label{sec:exp_set}

Our experiment consists of two parts: firstly, we benchmark commonly used datasets in the literature; secondly, we follow the tabular benchmark \cite{grinsztajn2022tree}, applying it to a broader range of datasets on both classification and regression tasks.

\paragraph{Datasets I.} In the first experimental setting, we evaluate BiSHop on 9 common classification datasets used in previous works \cite{grinsztajn2022tree,somepalli2021saint,gorishniy2021revisiting,huang2020tabtransformer}. 
These datasets vary in characteristics: some are well-balanced, and others show highly skewed class distributions; 
We set the train/validation/test proportion of each dataset as 70/10/20\%.
Please see \cref{sec:dataset_detail} for datasets' details.

\paragraph{Datasets II.} In the second experimental setting, 
we test BiSHop in the tabular benchmark \cite{grinsztajn2022tree}. 
The datasets compiled by this benchmark consist of 4 OpenML suites:
\begin{itemize}[ itemsep=0.1em]
    \item  Categorical Classification (\textbf{\texttt{CC}}, suite\_id: 334),
    \item  Numerical Classification (\textbf{\texttt{NC}}, suite\_id: 337), 

    \item  Categorical Regression (\textbf{\texttt{CR}}, suite\_id: 335),

    \item  Numerical Regression (\textbf{\texttt{NR}}, suite\_id: 336).
\end{itemize}
Both \textbf{\texttt{CC}} and \textbf{\texttt{CR}} include datasets with numerical and categorical features, whereas \textbf{\texttt{NC}} and \textbf{\texttt{NR}} only contain numerical features.
Due to limited computational resources, we randomly select one-third of the datasets from each suite for evaluation.
We evaluate BiSHop on each suit with 3-6 different datasets and truncate to 10,000 training samples for larger datasets (corresponding to medium-size regimes in the benchmark). 
For these datasets, we allocate 70\% of the data for the training set (7,000 samples). 
Of the remaining 30\%, we allocate 30\% for the validation set (900 samples), and the rest 70\% for the test set (2,100 samples). 
All samples are randomly chosen from the original datasets and undergo the identical preprocessing steps of the previous benchmark \cite{grinsztajn2022tree}.

\paragraph{Metrics.}
We use the AUC score for the 1st experimental setting, aligned with literature. 
For the 2nd experimental setting, we use accuracy for classification and the R$^2$ score for regression tasks, aligned with \citet{grinsztajn2022tree}.

\begin{table*}[h]
\centering
        \captionof{table}{\small
        \textbf{BiSHop versus SOTA Tabular Learning Methods (Dataset I).}
        We evaluate BiSHop against predominant SOTA methods, including deep learning methods (MLP, TabNet, TabTransformer, FT-Transformer, SAINT, TabPFN, TANGOS, T2G-FORMER) and tree-based methods (LightGBM, CatBoost, XGBoost), across various datasets.
        We report the average AUC scores (in \%) of 3 runs, with variances omitted as they are all $\le 1.3$\%.
        Results quoted from \cite{liu2022ptab,somepalli2021saint,borisov2022deep,huang2020tabtransformer} are marked with $^\star$, $^*$, $^\dagger$, and $^\ddag$, respectively.
        If multiple results are available across different benchmark papers, we quote the best one.
        When unavailable, we reproduce the baseline results independently.
        Hyperparameter optimization employs the ``sweep'' feature of Weights and Biases \cite{biewald2020experiment}, with 200 iterations of random search for each setting.
        Our results indicate that BiSHop outperforms both tree-based and deep-learning-based methods by a significant margin.}
        \vspace{0.5em}
        \label{tab:main_results}
                \resizebox{0.9\textwidth}{!}{%
        \begin{tabular}{lccccccccc}
\toprule
 & Adult & Bank & Blastchar & Income & SeismicBump & Shrutime & Spambase & Qsar & Jannis \\
\midrule
MLP & 72.5$^\ddag$ & 92.9$^\ddag$ & 83.9$^\ddag$ & 90.5$^\ddag$ & 73.5$^\ddag$ & 84.6$^\ddag$ & 98.4$^\ddag$ & 91.0$^\ddag$ & 82.59 \\
TabNet & 90.49 & 91.76$^*$ & 79.61$^*$ & 90.72$^*$  &  77.77 & 84.39 & 99.80 & 67.55$^*$  & 87.81 \\
TabTransformer & 73.7$^\ddag$ & 93.4$^\ddag$ & 83.5$^\ddag$ & 90.6$^\ddag$ & 75.1$^\ddag$ & 85.6$^\ddag$ & 98.5$^\ddag$ & 91.8$^\ddag$ & 82.85 \\
FT-Transformer & 90.60 & 91.83 & 86.06 & 92.15 & 74.60 & 80.83 & \textbf{100.00} & 92.04 & 89.02 \\

SAINT & 91.6$^\dagger$ & 93.30$^*$ & 84.67$^*$ & 91.67$^*$ & 76.6$^\star$ & 86.47$^*$ & 98.54$^*$ & 93.21$^*$ & 85.52 \\
TabPFN & 88.48  & 88.17 & 84.03 & 88.59  & 75.32  & 83.30 & 100 & 93.31 & 78.34 \\
TANGOS & 90.23  &  88.98 &  85.74 &  90.44 &  73.52 & 84.32  & 100   &  90.83 & 83.59 \\
T2G-FORMER &  85.96$^\diamond$ &  \textbf{94.47} &  85.40 &  92.35 &  82.58  &  86.42 & 100  &  94.86 & 73.68$^\diamond$ \\
\midrule
LightGBM & 92.9$^\dagger$ & 93.39$^*$ & 
83.17$^*$ & 92.57$^*$ & 77.43 & 85.36$^*$  & \textbf{100.00} & 92.97$^*$  & 87.48 \\
CatBoost & 92.8$^\dagger$  & 90.47$^*$ & 84.77$^*$ & 90.80$^*$ & 81.59 & 85.44$^*$  & \textbf{100.00} & 93.05$^*$  & 87.53 \\
XGBoost & 92.8$^\dagger$  & 92.96$^*$  & 81.78$^*$ & 92.31$^*$ & 75.3$^\star$ & 83.59$^*$  & \textbf{100.00} & 92.70$^*$  & 86.72 \\
\midrule
\rowcolor{LightCyan} \cellcolor{white}BiSHop & \textbf{92.97} & 93.95 & \textbf{88.49} & \textbf{92.97} & \textbf{91.88} & \textbf{87.99} & \textbf{100.00} & \textbf{96.14} & \textbf{90.63} \\

\bottomrule
\end{tabular}
        }
\end{table*}
\paragraph{Baselines I.}
In the first experimental setting, 
we select 5 deep learning and 3 tree-based baselines, including (i) DL-based method such as MLP, TabNet, TabTransformer, FT-Transformer \cite{gorishniy2021revisiting}, SAINT \cite{somepalli2021saint}, TabPNF \cite{hollmann2023tabpfn}, TANGOS \cite{jeffares2023tangos}, T2G-FORMER \cite{yan2023t2g}, and (ii) tree-based methods such as LightGBM, CatBoost, and XGBoost \cite{chen2015xgboost}.
For each dataset, we conduct up to 200 random searches on BiSHop to report the score of the best hyperparameter configuration.
We stop HPOs when observing the best result.
Baselines and benchmark datasets' results are quoted from competing papers when possible and reproduced otherwise.
We report the reproduced results in \cref{sec:sup_exp}.
Notably, we quote the best result from all baselines if multiple results are available.

\paragraph{Baselines II.} 
In the second experimental setting, we reference baselines results\footnote{\url{https://github.com/LeoGrin/tabular-benchmark}} from the benchmark paper \cite{gorishniy2021revisiting}, comprising 4 deep learning methods and 3 tree-based methods, including (i) DL-based method such as MLP, ResNet \cite{he2015deep}, FT-Transformer \cite{gorishniy2021revisiting}, SAINT \cite{somepalli2021saint} and (ii) tree-based methods such as RandomForest, GradientBoostingTree (GBDT), and XGBoost \cite{chen2015xgboost}.
We select the best results of each method from the benchmark \cite{grinsztajn2022tree}.
Notably, these best results use 400 HPOs according to \citet{grinsztajn2022tree}.
\begin{table*}[h]
\vspace{-1em}
\centering
\caption{\small
\textbf{BiSHop versus SOTA Tabular Learning Methods (Dataset II).}
Following the benchmark \cite{grinsztajn2022tree}, we evaluate BiSHop against SOTA methods, including deep learning methods (MLP, ResNet, FT-Transformer, SAINT) and tree-based methods (GBDT, RandomForest, XGBoost), across various datasets.
We randomly select a total of 19 datasets encompassing four different tasks: categorical classification (\textbf{\texttt{CC}}), numerical classification (\textbf{\texttt{NC}}), categorical regression (\textbf{\texttt{CR}}), and numerical regression (\textbf{\texttt{NR}}). 
\textbf{\texttt{CC}} and \textbf{\texttt{CR}} contain both categorical and numerical features, while \textbf{\texttt{NC}} and \textbf{\texttt{NR}} contain only numerical features.
Baseline results are quoted from the benchmark paper \cite{grinsztajn2022tree}.
We report with the best Accuracy scores for \textbf{\texttt{CC}} and \textbf{\texttt{NC}}, and R$^2$ score for \textbf{\texttt{CR}} and \textbf{\texttt{NR}}, (both in \%) obtained through HPO.
We also report the number of HPOs used in BiSHop.
Hyperparameter optimization of our method employs the ``sweep'' feature of Weights and Biases \cite{biewald2020experiment}. In the 19 different datasets, BiSHop delivers 11 optimal and 8 near-optimal results (within a 1.3\% margin), using less than 10\% (on average) of the number of HPOs used by the baselines.
}
\vspace{0.5em}

\resizebox{0.95\textwidth}{!}{%
\begin{tabular}{cgbbccccccc}
\toprule
 & \cellcolor{white}\footnotesize Dataset ID & \cellcolor{white}\footnotesize BiSHop  & \cellcolor{white}\footnotesize \# of HPOs &\footnotesize FT-Transformer &\footnotesize GBDT  &\footnotesize MLP &\footnotesize RandomForest &\footnotesize ResNet &\footnotesize  SAINT &\footnotesize   XGBoost\\
\midrule
 \multirow{3}{*}{\textbf{\texttt{CC}}} & 361282 & \textbf{66.08} & 16 & 65.63 & 65.76 & 65.32 & 65.53 & 65.23 & 65.52 & 65.70 \\ 
 & 361283 & \textbf{72.69} & 1  &  71.90 & 72.09 & 71.41 & 72.13 & 71.4 & 71.9 & 72.08 \\
& 361286 & \textbf{69.80} & 10 & 68.97 & 68.62 & 69.06 & 68.49 & 69.00 & 68.87 & 68.20 \\
\midrule
\multirow{5}{*}{\textbf{\texttt{CR}}} & 361093 & \textbf{98.98} & 23 & 98.06 & 98.34 & 98.07 & 98.25 & 98.04 & 97.77 & 98.42 \\ 
 & 361094 & 99.98 & 64  &  99.99 & \textbf{100} & 99.99 & \textbf{100} & 99.97 & 99.98 & \textbf{100} \\
& 361099 & 94.12 & 64 & 94.09 & 94.26 & 93.71 & 93.69 & 93.71 & 93.75 & \textbf{94.77} \\
& 361104 & 99.94 & 70 & 99.97 & \textbf{99.98} & \textbf{99.98} & \textbf{99.98} & 99.96 & 99.9 & \textbf{99.98} \\
& 361288 & 57.96 & 93 & 57.48 & 55.75 & 58.03 & 55.79 & \textbf{58.3} & 57.09 & 55.75 \\

\midrule
\multirow{5}{*}{\textbf{\texttt{NC}}} & 361055 & \textbf{78.29} & 4 & 77.73 & 77.52 & 77.41 & 76.35 & 77.53 & 77.41 &75.91 \\ 
& 361062 & \textbf{98.82} & 15 & 98.50 & 98.16 &94.70 & 98.24 & 95.22 & 98.21 & 98.35 \\
 & 361065 & \textbf{86.32} & 2 & 86.09 & 85.79 & 85.6 & 86.55 & 86.3 & 86.04 & 86.19 \\
& 361273 & \textbf{60.76} & 9 & 60.57 & 60.53 & 60.50 & 60.49 & 60.54 & 60.59 & 60.67 \\
& 361278 & \textbf{73.05} & 2 & 72.67 & 72.35 & 72.4 & 72.1 & 72.41 & 72.37 & 72.16 \\
\midrule
\multirow{6}{*}{\textbf{\texttt{NR}}} &        
361073 & \textbf{99.51} & 8  & \textbf{99.51} & 99.0 & 97.31 & 98.67 & 96.19 & \textbf{99.51} & 99.15 \\
& 361074 & 87.96 & 34 & 91.83 & 85.07 & 91.81 & 83.3 & 91.56 & \textbf{91.86} & 90.76 \\
& 361077 & 82.4 & 53 & 73.28 & \textbf{83.97} & 83.72 & 83.72 & 71.85 & 70.1 & 83.66 \\
 & 361079 & \textbf{60.76} & 19  &  53.09 & 57.45 & 48.62 & 50.16 & 51.77 & 46.79 & 55.42 \\
& 361081 & 98.67 & 13 & 99.69 & 99.65 & 99.52 & 99.31 & 99.67 & 99.38 & \textbf{99.76} \\
& 361280 & 56.98 & 96 & 57.48 & 54.87 & \textbf{58.46} & 55.27 & 57.81 & 56.84 & 55.49 \\
\midrule
\multirow{1}{*}{\textbf{Score}} & 
mean &  \textbf{81.21} & - & 80.34 & 80.48 & 80.3 & 79.84 & 79.81 & 79.68 & 80.65 \\
\midrule
\multirow{4}{*}{\textbf{Rank}} &        
mean & \textbf{2.79} & - & 3.58 & 4.21 & 4.74 & 5.53 & 5.05 & 5.26 & 3.84  \\
& min & \textbf{1} & - & 1 & 1 & 1 & 1 & 1 & 1 & 1 \\
& max & 8 & - & 6 & 8 & 8 & 8 & 8 & 8 & 8 \\
& med. & \textbf{1} & - & 4 & 4 & 5 & 6 & 5 & 5 & 3 \\

\bottomrule
\end{tabular}
}
\label{tab:main_results_pipeline}
\end{table*}

\paragraph{Setup.}
BiSHop's default parameter settings are as follows: Embedding dimension $G = 32$; Stride factor $L = 8$; Number of pooling vectors $C = 10$; Number of BiSHopModules $H = 3$; Number of aggregations in the encoder $r = 4$; Number of representations decoded $S = 24$; Dropout = 0.2; Learning rate $5 \times 10^{-5}$.
For numerical embedding, we only gather quantile information from the training data to process the embedding function.
For hyperparameter tuning, we use the ``sweep'' feature of Weights and Biases \cite{biewald2020experiment}.
Notably, due to computational constraints, we manually end the HPO once our method surpasses the best performance observed in the benchmarks.
We report the search space for all hyperparameters in \cref{tab:bihpo} and other training details in \cref{sec:baselines}.
We conduct the optimization on training/validation sets and report the average test set scores over 3 iterations, using the best-performing configurations on the validation set.
Implementation and training details are provided in the appendix.

\paragraph{Results.}
We summarize our results of the Baselines I in \cref{tab:main_results} and the results of the Baselines II in \cref{tab:main_results_pipeline}.
In \cref{tab:main_results}, BiSHop outperforms both tree-based and deep-learning-based methods by a significant margin in most datasets.
In \cref{tab:main_results_pipeline}, BiSHop achieves optimal or near-optimal results with fewer than 10\% numbers (on average) of HPO in a tabular benchmark \cite{grinsztajn2022tree}. 

\subsection{Ablation Studies}

We conduct the following sets of ablation studies on \textbf{Datasets I} align with \citet{grinsztajn2022tree}.
We include all details of ablation experiments in \cref{sec:additional_exp}.

\begin{figure}[h]
    \centering
        \includegraphics[width=.45\textwidth]{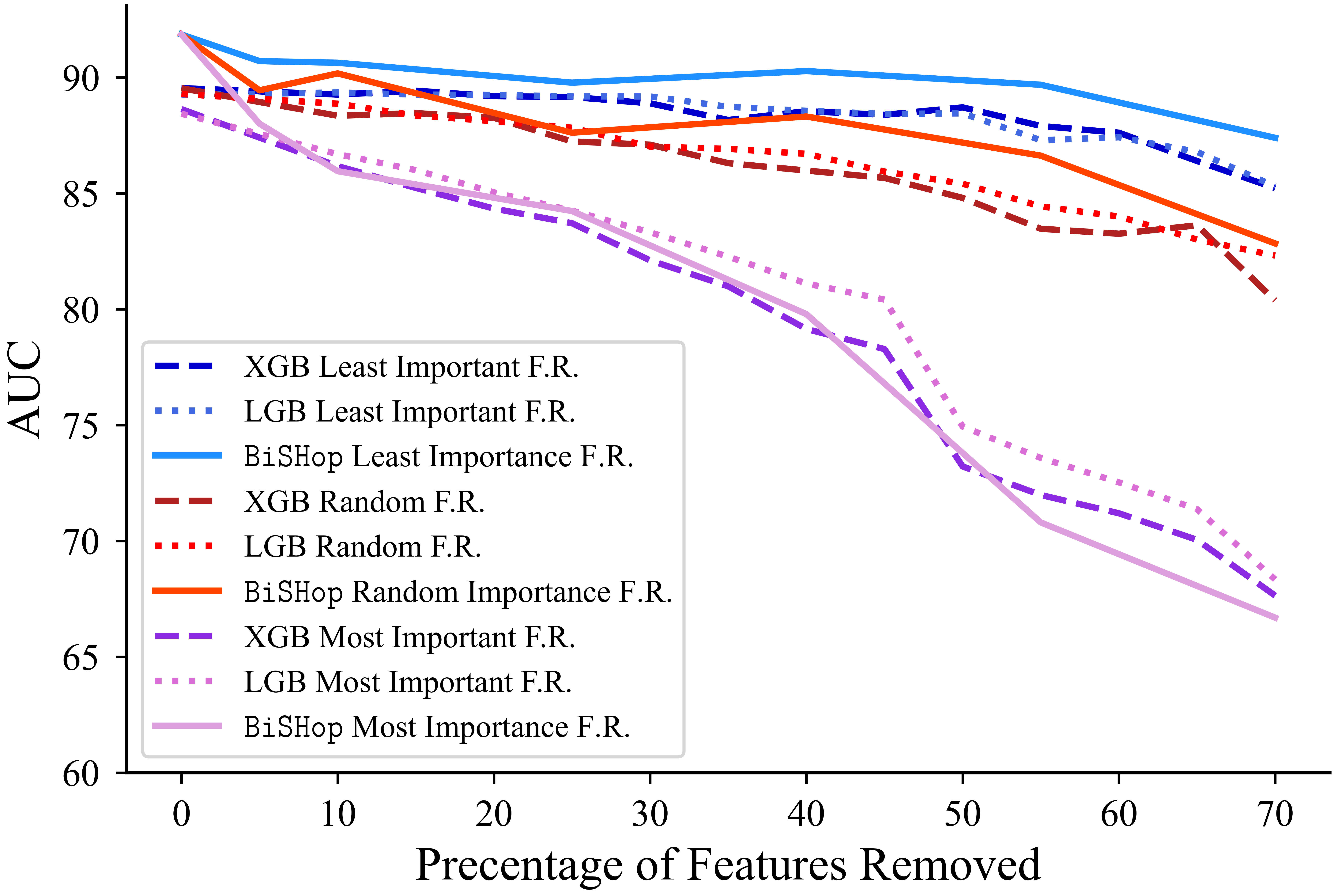}
        \captionof{figure}{\small
        \textbf{Changing Feature Sparsity.}
        Following \cite{grinsztajn2022tree}, we remove features in three ways: \textbf{randomly} (red), in \textbf{increasing} order of feature importance (purple), and in \textbf{decreasing} order of feature importance (blue), with feature importance determined by random forest.
        We report the average AUC score across all datasets for BiSHop, XGBoost, and LightGBM.
        The results highlight BiSHop's capability in handling sparse features.}
        \label{fig:feature_sparsity}
\end{figure}

\paragraph{Changing Feature Sparsity.}
    In \cref{fig:feature_sparsity}, we change feature sparsity on our datasets following \citet[Figure~4 \& 5]{grinsztajn2022tree}.
    Firstly, we compute the feature importance using Random Forest.
    Secondly, we remove
    features in both \textbf{increasing} (solid curves) and \textbf{decreasing} (dashed curves) order of feature importance.
    For each order, we report the average AUC score over all datasets at each percentage for BiSHop, XGBoost, and LighGBM.
    Our results indicate that BiSHop has the capacity to handle sparse features.

\paragraph{Rotation Invariance.}
In \cref{tab:rot}, we conduct experiments by rotating  the datasets and BiSHopModule's direction, both individual rotation  and combined rotation:
\begin{enumerate}[leftmargin=2.3em,itemsep=0.1em]
    \item [(R1)] Rotate the two Directions (Row- and Column-wise).
    \item [(R2)] Rotate the Datasets.
    \item [(R3)] Rotate the two Directions and the Datasets.
\end{enumerate}
Our results indicate (i) BiSHop is robust against column-row switch in BiSHopModule, and (ii) BiSHop addresses the Non-Rotationally Invariant Data Structure challenge \hyperref[item:C1]{(C1)}.

We report the average AUC score over all datasets for each rotation scenario. First, we assess (R1), where the results show a marginal ($<1\%$) performance drop across datasets. To further discuss the rotational invariance problem, we assess (R2) following the procedure outlined in \citet[Section 5.4]{grinsztajn2022tree}. The results for (R2) do not indicate a significant drop in performance. Additionally, the results for (R3) validate the findings from both (R1) and (R2). Our results confirm that BiSHop addresses \hyperref[item:C1]{(C1)}.

\paragraph{Hierarchy of BiSHopModule.}
In \cref{tab:stack}, we assess the impacts of stacking different layers of BiSHopModule.
We report the average AUC over \textbf{Datasets I} for different layers of BiSHopModule.
Our results indicate that 4 layers of BiSHopModule marginally maximize the performance.

\begin{table*}[ht]
\centering
\caption{\textbf{Component Ablation}. In the ablation study, we remove one component at a time. By evaluating different crucial components in BiSHop, we prove that each component contributes to various degrees of model performance. In particular, numerical embedding, decoder blocks, and the BiSHopModule contribute the most.}
\vspace{0.5em}
\resizebox{\textwidth}{!}{
\begin{tabular}{cbccccc}
\toprule
\cellcolor{white} Data & \cellcolor{white} BiSHop & w/o Cat Emb & w/o Num Emb & w/o Patch Emb & w/o Decoder & w/o BiSHopModule \\ 
\midrule
Adult        & \textbf{91.74}  & 90.91 & 89.40 & 91.32 & 88.18 & 91.28 \\
Bank         & \textbf{92.73}  & 90.88 & 77.21 & 91.14 & 91.93  & 91.98 \\
Blastchar    & 88.49  & 87.92 & \textbf{88.81} & 86.75 & 84.28  & 85.38 \\
Income       & \textbf{92.43}  & 91.01 & 90.38 & 91.56 & 91.44  & 91.36 \\
SeismicBumps & \textbf{91.42}  & 90.03 & 87.85 & 89.33 & 80.75  & 79.34 \\
Shrutime     & \textbf{87.38}  & 86.49 & 81.75 & 81.32 & 86.26  & 85.41 \\
Spambase     & \textbf{100}    & 100   & 100   & 100   & 100    & 100 \\
Qsar         & 92.85  & 91.15 & \textbf{94.69} & 91.50 & 93.04  & 91.65 \\
Jannis       & \textbf{89.66}  & 87.95 & 87.50 & 87.62 & 86.58  & 86.10 \\
\midrule
Average      & \textbf{91.86}  & 90.82 & 88.62 & 90.06 & 89.16  & 89.17 \\
\bottomrule
\end{tabular}
}
\label{tab:component}
\end{table*}

\paragraph{Component Analysis.}
In \cref{tab:component}, we remove each component one at a time. 
We report the implementation details in \cref{sec:ablation_comp}. 
For each removal, we report averaged AUC scores over all datasets.
Overall, each component contributes to varying degrees of performance improvement.

\paragraph{Comparison with the Dense Modern Hopfield Model.}
In \cref{sec:dense}, we compare the performance of Sparse Hopfield Models, Dense Hopfield Models, and Attention Mechanisms.
 Our results  that the generalized Sparse Hopfield Model outperforms the other two methods.
    
\paragraph{Convergence Analysis.}
In \cref{sec:convergence}, we compare the converging rate of Sparse and Dense Hopfield Models.
Our results indicate that the generalized sparse modern Hopfield model converges faster than the Dense Model.

\section{Conclusion}
\label{sec:conclusion}

We address the gap highlighted by \citet{grinsztajn2022tree} where deep learning methods trail behind tree-based methods. 
We present the Bi-Directional Sparse Hopfield Model (BiSHop) for deep tabular learning, inspired by the recent intersection of Hopfield models with attention mechanisms. 
Leveraging the generalized sparse Hopfield layers as its core component, BiSHop effectively handles the challenges of deep tabular learning, by incorporating two important inductive biases of tabular data (\hyperref[item:C1]{C1}, \hyperref[item:C2]{C2}).

\textbf{Comparing with Existing Works.}
Empirically, our model consistently surpasses SOTA tree-based and deep learning methods by 3\% across common benchmark datasets.
Moreover, our model achieves optimal or near-optimal results with only 16\% number of HPOs, compared with methods in the tabular benchmark  \cite{grinsztajn2022tree}.
We deem these results as closing the performance gap between DL-based and tree-based tabular learning methods, making BiSHop a promising solution for deep tabular learning.

\textbf{Limitation.}
One notable limitation of our study is the non-utilization of the external memory capabilities inherent in modern Hopfield models. We see the integration of these capabilities, especially in memory-augmented large models, as a compelling direction for future research.

\section*{Impact Statement}
\label{sec:broader}

Our work aim at addressing the long standing problem of tabular learning of DL-based model.
We do not expect any negative social impact of our work.

\section*{Acknowledgments}
JH would like to thank Dino Feng and Andrew Chen for enlightening discussions, the Red Maple Family for support, and Jiayi Wang for facilitating experimental deployments.
CX would like to thank Yibo Wen for helpful comments.
The authors would also like to thank the anonymous reviewers and program chairs for their constructive comments.

JH is partially supported by the Walter P. Murphy Fellowship.
HL is partially supported by NIH R01LM1372201, NSF CAREER1841569, DOE DE-AC02-07CH11359, DOE LAB 20-2261 and a NSF TRIPODS1740735.
H.-S.G. acknowledges support from the National Science and Technology Council, Taiwan under Grants No.~NSTC 113-2119-M-002 -021, No.~NSTC112-2119-M-002-014, No.~NSTC 111-2119-M-002-007, and No.~NSTC 111-2627-M-002-001, from the US Air Force Office of Scientific Research under Award Number FA2386-20-1-4052, and from the National Taiwan University under Grants No.~NTU-CC-112L893404 and No.~NTU-CC-113L891604. H.-S.G. is also grateful for the support from the ``Center for Advanced Computing and Imaging in Biomedicine (NTU-113L900702)'' through The Featured Areas Research Center Program within the framework of the Higher Education Sprout Project by the Ministry of Education (MOE), Taiwan, and the support from the Physics Division, National Center for Theoretical Sciences, Taiwan.
This research was supported in part through the computational resources and staff contributions provided for the Quest high performance computing facility at Northwestern University which is jointly supported by the Office of the Provost, the Office for Research, and Northwestern University Information Technology.
The content is solely the responsibility of the authors and does not necessarily represent the official
views of the funding agencies.

\clearpage
\newpage
\normalsize
\titlespacing*{\section}{0pt}{*1}{*1}
\titlespacing*{\subsection}{0pt}{*1.25}{*1.25}
\titlespacing*{\subsubsection}{0pt}{*1.5}{*1.5}

\setlength{\abovedisplayskip}{10pt}
\setlength{\abovedisplayshortskip}{10pt}
\setlength{\belowdisplayskip}{10pt}
\setlength{\belowdisplayshortskip}{10pt}

\onecolumn
\appendix
\part*{Supplementary Material}
\label{sec:append}

{
\setlength{\parskip}{-0em}
\startcontents[sections]
\printcontents[sections]{ }{1}{}
}
{\large \bf
\begin{itemize}
    \item \cref{sec:notation_tab}: Table of Notations

    \item \cref{sec:related_works}: Related Works

    \item \cref{sec:sup_theory}: Supplementary Theoretical Backgrounds

    \item \cref{sec:sup_exp}: Experimental Details

    \item \cref{sec:additional_exp}: Additional Numerical Experiments
\end{itemize}
}

\clearpage

\section{Table of Notations}
\label{sec:notation_tab}
\begin{table*}[ht]
    \centering
    \caption{Table of Notations.}
    \vspace{0.5em}
    \resizebox{ \textwidth}{!}{ 
    \begin{tabular}{ll}
    \toprule
    Notation & Description \\
    \midrule
    $\ba,\bb,\bc\ldots$ & Vectors
    \\
    $\bA,\bB,\bC\ldots$ & Matrices
    \\
    $\Braket{\ba,\bb}$ & Inner product of vectors $\ba$ and $\bb$, defined as $\ba^\sT \bb$ \\
    $[I]$ & Index set $\{1,\cdots,I\}$ for a positive integer $I$ \\
    $\norm{\cdot}_2$ & Spectral norm for matrices (aligned with $l_2$-norm for vectors) \\  
    \midrule
    $\bxi\in\R^d$ & Memory patterns (keys) \\
    $\bx\in\R^d$ & State/configuration/query pattern\\
    $\bm{\Xi}\coloneqq[\bxi_1,\cdots,\bxi_M]\in \R^{d\times M}$ & Shorthand for stored memory (key) patterns $\{\bxi_\mu\}_{\mu\in[M]}$ \\
    $n=\norm{\bx}$ & Norm of the query pattern \\
    $m = \Max_{\mu\in[M]}\norm{\bxi_\mu}$ & Maximum norm among the memory patterns 
    \\
    $\bm{\Xi}^\sT \bx$ & $M$-dimensional overlap vector  $\(\Braket{\bxi_1,\bx},\cdots,\Braket{\bxi_\mu,\bx},\cdots,\Braket{\bxi_M,\bx}\)$ in $\R^{M}$ \\
    \midrule
        $\[\bm{\Xi}^\sT \bx\]_\kappa$ & The $\kappa$-th element of  $\bm{\Xi}^\sT \bx$ 
        \\
        $\kappa$ & The number of non-zero element of $\Sparsemax$
        \\
        \midrule
        $n$ & Norm of $\bx$, denoted as $n\coloneqq\norm{\bx}$ \\
        $m$ & Largest norm of memory patterns, denoted as $m\coloneqq \Max_{\mu\in[M]}\norm{\bxi_\mu}$ 
        \\
        \midrule
        $R$ &
        The minimal Euclidean distance across all possible pairs of memory patterns, $R\coloneqq \half \Min_{\mu,\nu\in[M]}\norm{\bxi_\mu-\bxi_\nu}$
        \\
        $S_\mu$ &
        The sphere centered at the memory pattern $\bxi_\mu$ with finite radius $R$
        \\
        $\bx^\star_\mu$ & The fixed point of $\calT$ covered by $S_\mu$, i.e. $\bx_\mu^\star \in S_\mu$
        \\
        $\Delta_\mu$ & The separation of a memory pattern $\bxi_\mu$ from all other memory patterns $\bm{\Xi}$
        \\
        $\Tilde{\Delta}_\mu$&
        The separation of $\bxi_\mu$ at a given $\bx$ from all memory patterns $\bm{\Xi}$
        \\
    \midrule
    $ \bE( \cdot )$ & Embeddings \\
    $\bx \in \R^N$ & Single tabular data point with $N$ features (starting from Section \ref{sec:method}). \\
    $\mathtt{Concat} ([\bA, \bB ] ,\mathtt{axis}=0)$ & Concatenations of $\bA, \bB$ through first dimension ($\mathtt{axis}=1$ for concatenation through second dimension)\\
    $\bX $ & The internal embedding matrix of $\bx$. \\
    $\lceil \cdot \rceil$ & Ceiling function \\
    $\bX_{i,j}$ & The element of $i$-th rows and $j$-th columns in $\bX$ \\
    \bottomrule
    \end{tabular}
    }
    \label{tab:notations}
\end{table*}

\section{Related Works}
\label{sec:related_works}

\paragraph{Machine Learning for Tabular Data.}
Tabular data is a common data type across various domains such as time series prediction, fraud detection, physics, and recommendation systems. 
The state-of-the-art machine learning models for tabular data are tree-based models, particularly the family of gradient boosting decision trees (GBDT) \cite{chen2015xgboost, prokhorenkova2018catboost, ke2017lightgbm}.
In recent years, as deep learning model architectures have thrived in the natural language processing (NLP) and computer vision (CV) domains, many attempts have been made to adapt and apply these successful architectures, such as Multi-layer Perceptron (MLP) \cite{kadra2021well}, Convolutional Neural Networks (CNN) \cite{buturovic2020novel}, and Transformers \cite{huang2020tabtransformer, padhi2021tabular, somepalli2021saint}, to tabular data. 
Additionally, another line of work in deep learning involves creating differentiable tree-based models to enhance the capabilities of existing GBDT models \cite{arik2021tabnet, abutbul2020dnf, popov2019neural}.
However, unlike their dominance in NLP and CV, these deep learning models have struggled to surpass the performance of GBDTs on tabular data \cite{borisov2022deep, grinsztajn2022tree, ye2024closerlookdeeplearning}. 
Recent work, such as TabR \cite{gorishniy2023tabr}, shows some marginal advantages over GBDTs on certain datasets
For small datasets, TabPFN \cite{hollmann2023tabpfn}, utilizing Prior-Data Fitted Networks, performs better than tree-based methods. 
However, the memory and runtime usage scale quadratically with the training inputs.
T2G-FORMER \cite{yan2023t2g} fails to surpass XGBoost but performs better than other deep learning methods by learning feature relations. 
TANGOS \cite{jeffares2023tangos} narrows the gap between deep learning models and tree-based models by applying specific regularization techniques during neural network training.
To date, no deep learning model for tabular data has uniformly outperformed tree-based models

\paragraph{Modern Hopfield Models and Attention Mechanisms.}
Classical Hopfield models \cite{hopfield1984neurons,hopfield1982neural,krotov2016dense} are quintessential representations of the human brain's associative memory, primarily functioning to store and retrieve specific memory patterns.
Recently, there has been a resurgence of interest in associative memory models within the machine learning field \cite{burns2024semantically,burns2023simplicial,bietti2024birth,cabannes2024learning,cabannes2023scaling,Krotov2023,krotov2021hierarchical,krotov2020large,ramsauer2020hopfield}, attributed to advances in understanding memory storage capacities \cite{wu2024uniform,chaudhry2024long,demircigil2017model,krotov2016dense}, innovative architectures \cite{hoover2023energy,seidl2022improving,furst2022cloob,ramsauer2020hopfield}, and their biologically plausibility \cite{kozachkov2022building,krotov2020large}.
Notably, modern Hopfield models \cite{hu2024nonparametric,hu2024outlier,wu2024uniform,wu2023stanhop,hu2023sparse,ramsauer2020hopfield}\footnote{For an in-depth tutorial, see \cite{hopfeildblog2021}.} demonstrate not only a strong connection to transformer attention mechanisms in deep learning but also superior performance and a theoretically guaranteed exponential memory capacity
Their applicability spans diverse areas such as large language models \cite{hu2024outlier}, immunology \cite{widrich2020modern}, time series forecasting \cite{wu2023stanhop,auer2023conformal}, reinforcement learning \cite{paischer2022history}, and vision models \cite{furst2022cloob}.
In this context, this work emphasizes refining this line of research towards sparser models. 
We posit that this effort is crucial in guiding future research towards Hopfield-driven design paradigms and bio-inspired computing systems.

\paragraph{Sparse Attention.}
The attention mechanisms of transformers have demonstrated unparalleled performance in many domains, such as large language models \cite{xu2024do,lwd+23,zhang2024h2o,chowdhery2022palm,brown2020language}, time series prediction \cite{zhou2022film,zhou2021informer}, and biomedical science \cite{zhou2023dnabert,yang2021identifying,ji2021dnabert}.
However, the standard transformer architecture relies heavily on a dense quadratic attention score matrix. This structure presents computational challenges, especially for longer sequences, given its $\calO(n^2)$ complexity for an input sequence of length $n$.
In response to this challenge, a wealth of research has introduced sparse variants of attention mechanisms and transformers, aiming to strike a balance between computational efficiency and model expressiveness. For a comprehensive review, readers may refer to \cite{tay2022efficient}. Generally, these sparse attention/transformer methodologies fall into two categories:
\begin{enumerate}
\setlength\itemsep{0em}
    \item \textbf{Structured-Sparsity Attentions} \cite{beltagy2020longformer, qiu2019blockwise, child2019generating}: These methods utilize structured, predetermined patterns in the attention matrix. Typically, each sequence token attends to a predetermined subset of tokens instead of the entire sequence.
    \item \textbf{Dynamic Sparsity via Normalization Maps} \cite{peters2019sparse,correia2019adaptively,krotov2016dense}: In contrast to structured sparsity, these methods dynamically determine sparsity, centering on the most relevant input elements. Despite potentially retaining a space complexity of $\calO(n^2)$, they dynamically tailor sparsity patterns to the data, bolstering scalability and clarity.
\end{enumerate}
This work aligns more closely with the second category, as we employ the generalized sparse modern Hopfield model \cite{wu2023stanhop} by substituting the softmax function in the modern Hopfield models with \textit{sparsity-inducing alternatives}. 
A theoretical analysis of the efficiency of modern Hopfield models can be found in \cite{hu2024computational}.

\section{Supplementary Theoretical Backgrounds}
\label{sec:sup_theory}
To highlight the computational benefits of the generalized sparse {modern} Hopfield model, we quote relevant results from \cite{wu2023stanhop} here.

\subsection{Definition of Memory Storage and Retrieval and Separation of Patterns}

We adopt the formal definition of memory storage and retrieval from \cite{ramsauer2020hopfield} for continuous patterns.
\begin{definition} [Stored and Retrieved]
\label{def:stored_and_retrieved}
Assuming that every pattern $\bxi_\mu$ is surrounded by a sphere $S_\mu$ with finite radius $R\coloneqq \half \Min_{\mu,\nu\in[M]}\norm{\bxi_\mu-\bxi_\nu}$, we say $\bxi_\mu$ is \textit{stored} if there exists a generalized fixed point of $\calT$, $\bx^\star_\mu \in S_\mu$, to which all limit points $\bx \in S_\mu$ converge to, and $S_\mu \cap S_\nu=\emptyset$ for $\mu \neq \nu$. 
We say $\bxi_\mu$ is $\epsilon$-\textit{retrieved} by $\calT$ with $\bx$ for an error.
\end{definition}

We then introduce the definition of pattern separation for later convenience.
\begin{definition}[Pattern Separation]
\label{def:pattern_separation}
Let's consider a memory pattern $\bxi_\mu$ within a set of memory patterns $\bm{\Xi}$.
\begin{enumerate}
    \item The separation metric $\Delta_\mu$ for $\bxi_\mu$ with respect to other memory patterns is the difference between its self-inner product and the maximum inner product with any other pattern
    \begin{align}
    \Delta_\mu = \Braket{\bxi_\mu,\bxi_\mu} - \Max_{\nu,\nu\neq \mu}\Braket{\bxi_\mu,\bxi_\nu}.
    \label{eqn:sep_delta_mu}
    \end{align}
    \item Given a specific pattern $\bx$, the relative separation metric $\Tilde{\Delta}_\mu$ for $\bxi_\mu$ with respect to other patterns in $\bm{\Xi}$ is defined as:
    \begin{align}
    \Tilde{\Delta}_\mu = \Min_{\nu,\nu\neq \mu} \left( \Braket{\bx,\bxi_\mu} - \Braket{\bx,\bxi_\nu} \right).
    \label{eqn:tilde_sep_delta_mu}
    \end{align}
\end{enumerate}
\end{definition}

\subsection{Supplementary Theoretical Results for Generalized Sparse Modern Hopfield Model}
\label{sec:theoretical_properties}

\begin{theorem}[Retrieval Error, Theorem~3.1 of \cite{wu2023stanhop}]
\label{thm:eps_sparse_dense}
Let $\calT_{\text{Dense}}$ be the retrieval dynamics of the dense modern Hopfield model \cite{ramsauer2020hopfield}.
It holds
$
\norm{\calT(\bx)-\bxi_\mu} 
\leq
\norm{\calT_{\text{Dense}}(\bx)-\bxi_\mu}
$
for all $\mu$.
\end{theorem}
\noindent
\cref{thm:eps_sparse_dense} implies two computational advantages:
\begin{corollary}[Faster Convergence]
\label{coro:faster_convergence}
Computationally, \cref{thm:eps_sparse_dense} suggests that $\calT$ converges to fixed points using fewer iterations than $\calT_{\text{dense}}$ for the same error tolerance. 
This means that $\calT$ retrieves stored memory patterns more quickly and efficiently than its dense counterpart.
\end{corollary}
\begin{corollary}[Noise-Robustness]
\label{coro:noise}
In cases of noisy patterns with noise $\bm{\eta}$, i.e. $\Tilde{\bx}=\bx+\bm{\eta}$ (noise in query) or $\Tilde{\bxi}_\mu=\bxi_\mu+\bm{\eta}$ (noise in memory), the impact of noise $\bm{\eta}$ on the sparse retrieval error $\norm{\calT(\bx)-\bxi_\mu}$ is linear for  $\alpha\geq2$, while its effect on the dense retrieval error $\norm{\calT_{\text{Dense}}(\bx)-\bxi_\mu}$ (or $\norm{\calT(\bx)-\bxi_\mu}$ with $2\geq\alpha\geq1$) is exponential.
\end{corollary}
\begin{remark}
   \cref{coro:faster_convergence} does not imply computational efficiency.
   The  proposed model's sparsity falls under the category of \textit{sparsity-inducing normalization maps} \cite{tay2022efficient,peters2019sparse,correia2019adaptively,krotov2016dense}.
   This means that, during the forward pass, the space complexity remains at $\calO(n^2)$, on par with the dense modern Hopfield model. 
\end{remark}
\begin{remark}
    Nevertheless, \cref{coro:faster_convergence}  suggests a specific type of ``efficiency" related to faster memory retrieval compared to the dense Hopfield model.
    In essence, a retrieval dynamic with a smaller error converges faster to the fixed points (stored memories), thereby enhancing efficiency.
\end{remark}

\clearpage
\section{Experimental Details}
\label{sec:sup_exp}
\paragraph{Computational Hardware.}
All experiments are conducted on the platforms equipped with NVIDIA GEFORCE RTX 2080 Ti,  Tesla A100 SXM GPU, and INTEL XEON SILVER 4214 @ 2.20GHz.

\subsection{Additional Details on Datasets}
\label{sec:dataset_detail}
We describe all the datasets used in our experiments in \cref{tab:datasetsinfo} and provide the download links to each dataset in \cref{tab:link}.

\begin{table*}[h]
\centering
\caption{Dataset Sources.}
\vspace{0.5em}
\begin{tabular}{ll}
\toprule
Dataset & URL 
\\
\midrule
Adult & \url{http://automl.chalearn.org/data} 
\\
Bank & \url{https://archive.ics.uci.edu/ml/datasets/bank+marketing} 
\\
Blastchar & \url{https://www.kaggle.com/blastchar/telco-customer-churn} 
\\
Income & \url{https://www.kaggle.com/lodetomasi1995/income-classification} 
\\
SeismicBumps & \url{https://archive.ics.uci.edu/ml/datasets/seismic-bumps} 
\\
Shrutime & \url{https://www.kaggle.com/shrutimechlearn/churn-modelling} 
\\
Spambase & \url{https://archive.ics.uci.edu/ml/datasets/Spambase} 
\\
Qsar & \url{https://archive.ics.uci.edu/dataset/254/qsar+biodegradation} 
\\
Jannis & \url{http://automl.chalearn.org/data} 
\\
\bottomrule
\end{tabular}
\label{tab:link}
\end{table*}

\begin{table*}[h]
\centering
\caption{\textbf{Details of Datasets.} 
We summarize the statistics of {9} datasets we have used in Baseline I, {8} of which involve binary classification and 1 of which involve multi-class classification (4 classes).}
\vspace{0.5em}
\resizebox{ \textwidth}{!}{%
\begin{tabular}{lccccccccc}
\toprule
 & Adult & Bank & Blastchar & Income & SeismicBump & Shrutime & Spambase & Qsar & Jannis \\
\midrule

\# Numerical & 6 & 7 & 3 & 6 & 14 & 6 & 58 & 41 & 54 \\
\# Categorical & 8 & 9 & 16 & 8 & 4 & 4 & 0 & 0 & 0 \\
\# Train & 34190 & 31648 & 4923 & 34189 & 1809 & 7001 & 3221 & 738 & 58613 \\
\# Validation & 9769 & 9042 & 1407 & 9768 & 517 & 2000 & 920 & 211 & 16747\\
\# Test & 4884 & 4522 & 703 & 4885 & 258 & 1000 & 461 & 106 & 8373 \\
\# Task type &  Bi-Class & Bi-Class & Bi-Class & Bi-Class & Bi-Class & Bi-Class & Bi-Class & Bi-Class & Multi-Class\\
\bottomrule
\end{tabular}
}
\label{tab:datasetsinfo}
\end{table*}

\begin{table*}[ht]

\centering
\caption{\textbf{Details of Datasets.} 
We summarize the statistics of {19} datasets covering four suite: categorical classification (\textbf{\texttt{CC}}), numerical classification (\textbf{\texttt{NC}}), categorical regression (\textbf{\texttt{CR}}), and numerical regression (\textbf{\texttt{NR}}).}
\vspace{0.5em}

\resizebox{.8\textwidth}{!}{%
\begin{tabular}{clcccl}
\toprule
 & Dataset ID & Dataset Name   & \# of Categorical & \# of Numerical \\
\midrule
 \multirow{3}{*}{\textbf{\texttt{CC}}} & 361282 & albert  & 11 & 21 \\ 
 & 361283 & default-of-credit-card-clients   &  2 & 20 \\
& 361286 & compas-two-years  & 9 & 3 \\
\midrule
\multirow{5}{*}{\textbf{\texttt{CR}}} & 361093 & analcatdata\_supreme  & 5 & 3 \\ 
 & 361094 & visualizing\_soil   &  1 & 4 \\
& 361099 & Bike\_Sharing\_Demand  & 5 & 7 \\
& 361104 & SGEMM\_GPU\_kernel\_performance  & 6 & 4 \\
& 361288 & abalone  & 1 & 8 \\

\midrule
\multirow{5}{*}{\textbf{\texttt{NC}}} & 361055 & credit  & 0 & 10 \\ 
& 361062 & pol  & 0 & 26 \\
 & 361065 & MagicTelescope  & 0 & 10 \\
& 361273 & Diabetes130US  & 0 & 7  \\
& 361278 & heloc  & 0 & 22 \\
\midrule
\multirow{6}{*}{\textbf{\texttt{NR}}} &        
361073 & pol   & 0 & 27 \\
& 361074 & elevators  & 0 & 17 \\
& 361077 & Ailerons  & 0 & 34 \\
 & 361079 & house\_16H   &  0 & 17  \\
& 361081 & Brazilian\_houses  & 0 & 9  \\
& 361280 & abalone  & 0 & 8  \\
\bottomrule
\end{tabular}
}
\label{tab:openml_meta}
\end{table*}

The links to the four OpenML suites from \cite{grinsztajn2022tree} are \textbf{\texttt{CC}}: \footnote{\url{https://www.openml.org/search?type=benchmark&sort=date&study_type=task&id=300}}, \textbf{\texttt{NC}}\footnote{\url{https://www.openml.org/search?type=benchmark&study_type=task&sort=tasks_included&id=298}},
\textbf{\texttt{CR}}\footnote{\url{https://www.openml.org/search?type=benchmark&study_type=task&sort=tasks_included&id=299}},
\textbf{\texttt{NR}}\footnote{\url{https://www.openml.org/search?type=benchmark&study_type=task&sort=tasks_included&id=297}}

\subsection{Baselines}
\label{sec:baselines}
We evaluate BiSHop by comparing it to SOTA tabular learning methods, specifically selecting top performers in recent studies \cite{grinsztajn2022tree,somepalli2021saint,gorishniy2021revisiting}.

\begin{itemize}
    \item \textbf{LightGBM} \cite{ke2017lightgbm}.
    \item \textbf{CatBoost} \cite{prokhorenkova2018catboost}.
    \item \textbf{XGBoost}
    \cite{chen2015xgboost}.
    \item \textbf{MLP}
    \cite{somepalli2021saint}.
    \item \textbf{TabNet}
    \cite{arik2021tabnet}.
    \item \textbf{TabTransformer}
    \cite{huang2020tabtransformer}.
    \item \textbf{FT-Transformer}
    \cite{gorishniy2021revisiting}.
    \item \textbf{SAINT}
    \cite{somepalli2021saint}.
    \item \textbf{TabPFN}
    \cite{hollmann2023tabpfn}. We implement TabPFN using 32 data permutations for ensemble, as in the original paper setting and truncate the training set to 1024 instances.
    \item \textbf{T2G-FORMER}
    \cite{yan2023t2g}. We implement T2G-FORMER by applying quantile transformation from the Scikit-learn library to Baseline I datsets, aligning with the default setting in. The hyperparameter space is at \cref{table:t2g}.
    \item \textbf{TANGOS}
    \cite{jeffares2023tangos}. We adapted the official TANGOS source code to include the datasets from Baseline I alongside the original datasets.  The hyperparameter space is at \cref{table:HPO_tangos}.
\end{itemize}

\textbf{Selection of Benchmark. }
We select \citet{grinsztajn2022tree} as our benchmark for several reasons. 
Unlike other benchmarks that focus solely on tasks such as classification \cite{gardner2023benchmarking}, this benchmark encompasses both regression and classification tasks. Additionally, it provides results from 400 hyperparameter optimization (HPO) trials, ensuring a thorough hyperparameter search for each model.
In contrast, some methods, such as \cite{mcelfresh2023neural}, restrict HPO to 10 hours on a specific GPU, which can be insufficient for deep-learning-based methods like BiSHop that require more training time compared to tree-based methods. Moreover, comparing models under the same time constraints on different GPUs is inherently unfair.

\subsection{Implementation Details}
\paragraph{Data Prepossessing.}
We initially transform the categorical features into discrete values (e.g., 0, 1, 2, 3) and retain the raw numerical features without any processing. 
For tree-based method baselines, we employ the built-in categorical embedding method for categorical features. 
For deep learning baselines, we further encode the categorical features using one-hot encoding.

\paragraph{Evaluation.}
For each model's hyperparameter configuration, we run 3 experiments using the best configuration and report the average AUC score on the test set.

\subsection{Training Details}
\begin{table*}[h]
\centering
\caption{BiSHop Hyperparameter Space.}
\vspace{0.5em}
\resizebox{\textwidth}{!}{
\begin{tabular}{lccccccccc}
\toprule
 & Parameter & Distribution & {Default} \\
\midrule
& Number of representation
decoded & [2, 4, 8, 16, 24, 32, 48, 64, 128, 256, 320]& {24}  \\
& Stride factor & [1, 2, 4, 6, 8, 12, 16, 24] & {8} \\
& Embedding dimension & [16, 24, 32, 48, 64, 128, 256, 320] & {32} \\
& Number of aggregation in encoder & [2, 3, 4, 5, 6, 7, 8]& {4}  \\
& Number of pooling vector & [5, 10, 15] & {10}  \\
& Dimension of hidden layers ($D^\text{model}$) & [64, 128, 256, 512, 1024]& {512}  \\
& Dimension of feedforward network (in MLP) & [128, 256, 512, 1024]& {256}  \\
& Number of multi-head attention & [2, 4, 6, 8, 10, 12]& {4}  \\
& Number of Encoder & [2, 3, 4, 5]& {2} \\
& Number of Decoder & [0, 1]& {2}  \\
& Learning rate & LogUniform[(1e-6, 1e-4)& {5e-5}  \\
& ReduceLROnPlateau & factor=0.1, eps=1e-6& {factor=0.1, eps=1e-6}  \\
\bottomrule
\end{tabular}
}
\label{tab:bihpo}
\end{table*}

\paragraph{Learning Rate Scheduler.}
We use $\mathtt{ReduceLROnPlateau}$ to fine-tune the learning rate to improve convergence and model training progress.

\paragraph{Optimizer.} 
We use the Adam optimizer to minimize cross-entropy. The coefficients of the Adam optimizer, betas, are set to (0.9, 0.999).

\paragraph{Patience.}
We continue training until there are $\mathtt{Patience=20}$ consecutive epochs where the validation loss doesn't decrease, or we reach 200 epochs. Finally, we evaluate our model on the test set with the last checkpoint.

\paragraph{HPO.}
We report the number of HPO for each dataset from baseline I in \cref{tab:baseline1hpo}. 
We report hyperparameter configurations for CatBoost in \cref{table:HPO_catboost}, LightGBM in \cref{table:HPO_lightgbm}, TabNet in \cref{table:HPO_tabnet}, XGBoost in \cref{table:HPO_xgboost}, T2G-Former in \cref{table:t2g}, Tangos in \cref{table:HPO_tangos}.
We follow the same procedure of HPOs for Tangos and T2G-Former in \citet{yan2023t2g} and \citet{jeffares2023tangos}, including the number of trials.
For other methods, we follow the same settings as BiSHop.

{\paragraph{Hyperparameter Importance Analysis.}
During random hyperparameter search, we observe that learning rate is the most important hyperparameter (see \cref{table:hyperparameter_importance}). 
We use WandB "sweep" features \cite{biewald2020experiment} to calculate the importance of each hyperparameter. 
Our findings agree with \cite{grinsztajn2022tree} suggesting that learning rate is the most important hyperparameter for both neural network and gradient-boosted trees.}
\begin{table*}[h]
\centering
\caption{\textbf{Hyperparameter Importance Scores.}
The importance is calculate from features importance in RandomForest, averaging across all datasets. This results highlight learning rate is the most crucial hyperparameter.} 
\vspace{0.5em}
\label{table:hyperparameter_importance}
\begin{tabular}{lc}
\hline
Hyperparameter & RF Importance \\
\hline
Learning rate & 0.17 \\
Dropout & 0.10 \\
Number of heads &  0.08 \\
Number of aggregation & 0.06 \\
Dimension of hidden layers & 0.06 \\
Dimension of feed-forward network & 0.13 \\
Number of pooling factor & 0.05 \\
Number of encoder layer & 0.05 \\
Number of representation decoded & 0.10 \\
\hline
\end{tabular}
\end{table*}

\begin{table*}[h]
\centering
\caption{{Number of HPO in Baseline I.}}
\vspace{0.5em}
\begin{tabular}{lc}
\toprule
Dataset & \# of HPO \\
\midrule
Adult & 36 \\
Bank & 26\\
Blastchar & 52\\
Income &  174\\
SeismicBumps & 200\\
Shrutime & 16\\
Spambase & 1\\
Qsar & 67\\
Jannis & 137\\
\bottomrule
\end{tabular}
\label{tab:baseline1hpo}
\end{table*}

\begin{table*}[ht]
\centering
\caption{Hyperparameter Configurations for CatBoost.}
\vspace{0.5em}
\begin{tabular}{lccc}
\toprule
{Parameter} & {Distribution} & Default \\
\midrule
Depth & UniformInt[3,10] & 6 \\
L2 regularization coefficient & UniformInt[1,10] & 3 \\
Bagging temperature & Uniform[0,1] & 1 \\
Leaf estimation iterations & UniformInt[1,10] & None \\
Learning rate & LogUniform[1e-5, 1] & 0.03 \\
\bottomrule
\end{tabular}

\label{table:HPO_catboost}
\end{table*}

\begin{table*}[ht]
\centering
\caption{Hyperparameter Configurations for LightGBM.}
\vspace{0.5em}
\begin{tabular}{lccc}
\toprule
{Parameter} & {Distribution} & Default \\
\midrule
Number of estimators & [50, 75, 100, 125, 150] & 100 \\
Number of leavs & UniformInt[10, 50] & 31 \\
Subsample & UniformInt[0, 1] & 1 \\
Colsample & UniformInt[0, 1] & 1 \\
Learning rate & LogUniform[1e-1,1e-3] & None \\
\bottomrule
\label{table:HPO_lightgbm}
\end{tabular}
\end{table*}

\begin{table*}[ht]
\centering
\caption{Hyperparameter Configurations for TabNet.}
\vspace{0.5em}
\begin{tabular}{lccc}
\toprule
{Parameter} & {Distribution} & Default \\
\midrule
n\_d & UniformInt[8,64] & 8 \\
n\_a & UniformInt[8,64] & 8 \\
n\_steps & UniformInt[3,10]& 3 \\
Gamma & Uniform[1.0,2.0] & 1.3 \\
n\_independent & UniformInt[1,5] & 2 \\
Learning rate & LogUniform[1e-3, 1e-1] & None \\
Lambda sparse & LogUniform[1e-4, 1e-1] & 1e-3 \\
Mask type & entmax & sparsemax\\
\bottomrule
\label{table:HPO_tabnet}
\end{tabular}
\end{table*}

\begin{table*}[ht]
\centering
\caption{Hyperparameter Configurations for XGBoost.}
\vspace{0.5em}
\begin{tabular}{lccc}
\toprule
{Parameter} & {Distribution} & Default \\
\midrule
Max depth & UniformInt[3,10] & 6 \\
Minimum child weight & LogUniform[1e-4,1e2] & 1 \\
Subsample & Uniform[0.5,1.0] & 1 \\
Learning rate & LogUniform[1e-3,1e0] & None \\
Colsample bylevel & Uniform[0.5,1.0] & 1 \\
Colsample bytree & Uniform[0.5,1.0] & 1 \\
Gamma & LogUniform[1e-3,1e2] & 0 \\
Alpha & LogUniform[1e-1,1e2] & 0 \\
\bottomrule
\end{tabular}
\label{table:HPO_xgboost}
\end{table*}

\begin{table*}[ht]
\centering
\caption{Hyperparameter Configurations for T2G-FORMER.}
\vspace{0.5em}
\begin{tabular}{lccc}
\toprule
{Parameter} & {Distribution} & Default \\
\midrule
\# layers & UniformInt[1,3] & None \\
Feature embedding size & UniformInt[64,512] & None \\
Residual Dropout & Const(0.0) & None \\
Attention Dropout & Uniform[0, 0.5] & None \\
FNN Dropout & Uniform[0, 0.5] & None \\
Learning rate (main backbone) & LogUniform[3e-5, 3e-4] & None \\
Learning rate (column embedding) & LogUniform[5e-3, 5e-2] & None \\
Weight decay & LogUniform[1e-6, 1e-3] & None \\
\bottomrule
\end{tabular}
\label{table:t2g}
\end{table*}

\begin{table*}[ht]
\centering
\caption{Hyperparameter Configurations for TANGOS.}
\vspace{0.5em}
\begin{tabular}{lccc}
\toprule
{Parameter} & {Distribution} & Default \\
\midrule
$\lambda_1$ & LogUniform[0.001,10] & None \\
$\lambda_2$ & LogUniform[0.0001,1] & None \\
\bottomrule
\end{tabular}

\label{table:HPO_tangos}
\end{table*}

\begin{table*}[h]
\centering
\caption{BiSHop Hyperparameter Search Space.}
\vspace{0.5em}
\resizebox{\textwidth}{!}{
\begin{tabular}{@{}ccccccc@{}}
\toprule
Data & BiSHop & w/o Cat Emb & w/o Num Emb & w/o Patch Emb & w/o Decoder & w/o GSH \\ 
\midrule
Adult        & 91.50  & 91.54 & 89.40 & 91.32 & 0.306  & 0.306 \\
Bank         & 92.23  & 92.90 & 77.21 & 91.14 & 0.346  & 0.306 \\
Blastchar    & 88.49  & 88.05 & 88.81 & 86.75 & 0.434  & 0.306 \\
Income       & 91.47  & 91.38 & 90.38 & 91.56 & 0.462  & 0.306 \\
SeismicBumps & 91.42  & 91.72 & 87.85 & 89.33 & 0.524  & 0.306 \\
Shrutime     & 87.38  & 87.16 & 81.75 & 81.32 & 0.524  & 0.306 \\
Spambase     & 100    & 100   & 100   & 100   & 0.524  & 0.306 \\
Qsar         & 92.85  & 91.54 & 94.69 & 91.50 & 0.524  & 0.306 \\
Jannis       & 88.27  & 87.95 & 87.50 & 0.610 & 0.524  & 0.306 \\
\midrule
Average      & 91.51  & 91.36 & 87.50 & 0.610 & 0.524  & 0.306 \\
\bottomrule
\end{tabular}
}
\end{table*}

\clearpage
\section{Additional Numerical Experiments}
\label{sec:additional_exp}

\subsection{Component Analysis}
\label{sec:ablation_comp}

We separately remove each component of BiSHop to assess its impact on performance. We use the default hyperparameters as specified in \cref{tab:bihpo} for the remaining components. We report the average AUC score of three runs using the default parameters for all datasets in \cref{tab:component}. 
\begin{itemize}
    \item Without Cat Emb: We remove both individual and shared embedding methods as described in the tabular embedding section, replacing them with PyTorch's embedding layers (\texttt{torch.nn.Embedding}) while keeping the embedding dimension unchanged.
    \item Without Num Emb: We remove the \texttt{Piecewise Linear Encoding} method for numerical features, directly concatenating numerical features with the output of categorical embedding {as detailed in \cref{Categorical Embedding}}.
    \item Without Patch Embedding: We remove the patch embedding method by setting the stride factor $L$ to 1.
    \item Without Decoder: We remove the decoder blocks {in BiSHop} and pass the encoded data directly to MLP predictor.
    \item Without BiSHopModule: We replace the column-wise block and row-wise block in the BiSHop module with a {MLP of hidden size 512}.
\end{itemize}
The results demonstrate that each component contributes to varying degrees to the BiSHop model, with numerical embedding, decoder blocks, and the BiSHopModule being the most significant contributors.

\clearpage
\subsection{Comparison with the Dense Modern Hopfield Model}
\label{sec:dense}

Using the default hyperparameters of BiSHop, we evaluate its performance with three distinct layers: (i) \texttt{GSH} (generalized sparse modern Hopfield model), (ii) \texttt{Hopfield} (dense modern Hopfield model \cite{ramsauer2020hopfield}), and (iii) \texttt{Attn} (attention mechanism \cite{vaswani2017attention}).
We report the average AUC score over 10 runs in \cref{tab:abl_main}.
\begin{table*}[h]
\centering
\caption{\small
\textbf{
Comparing the Performance of Sparse versus Dense Modern Hopfield Models and Attention Mechanism.}
We contrast the performance of our generalized sparse modern Hopfield model with that of the dense modern Hopfield model and the attention mechanism. 
We achieve this by substituting the $\mathtt{GSH}$ layer with the $\mathtt{Hopfield}$ layer from \cite{ramsauer2020hopfield} and the $\mathtt{Attn}$ layer from \cite{vaswani2017attention}. 
We report the average AUC score (in \%) over 10 runs, with variances omitted as they are all $\le 1.1$\%. 
The results indicates the superior performance of our proposed generalized sparse modern Hopfield model across datasets.
}
\vspace{0.5em}
\resizebox{ \textwidth}{!}{%
\begin{tabular}{lcccccccccc}
\toprule
AUC (\%) & Adult & Bank & Blastchar & Income & SeismicBump & Shrutime & Spambase & Qsar & Jannis & \textbf{Mean AUC} \\
\midrule
$\mathtt{GSH}$ & \textbf{91.74} & \textbf{92.73} & \textbf{88.49} & \textbf{92.43} & \textbf{91.42} & \textbf{87.38} & \textbf{100} & \textbf{92.85} & \textbf{89.66} & \textbf{91.86} \\
$\mathtt{Hopfield}$ & 91.72 & 92.60 & 85.31 & 91.65 & 78.63 & 86.81 & 100 & 91.27 & 85.04 & 89.23 \\
$\mathtt{Attn}$ & 91.44 & 92.46 & 83.14 & 91.46 & 78.42 & 83.04 & 100 & 89.88 & 88.28 & 88.68  \\
\bottomrule
\label{tab:abl_main}
\end{tabular}
}
\end{table*}

\begin{figure*}[h]
    \centering
    \includegraphics[width=\textwidth]{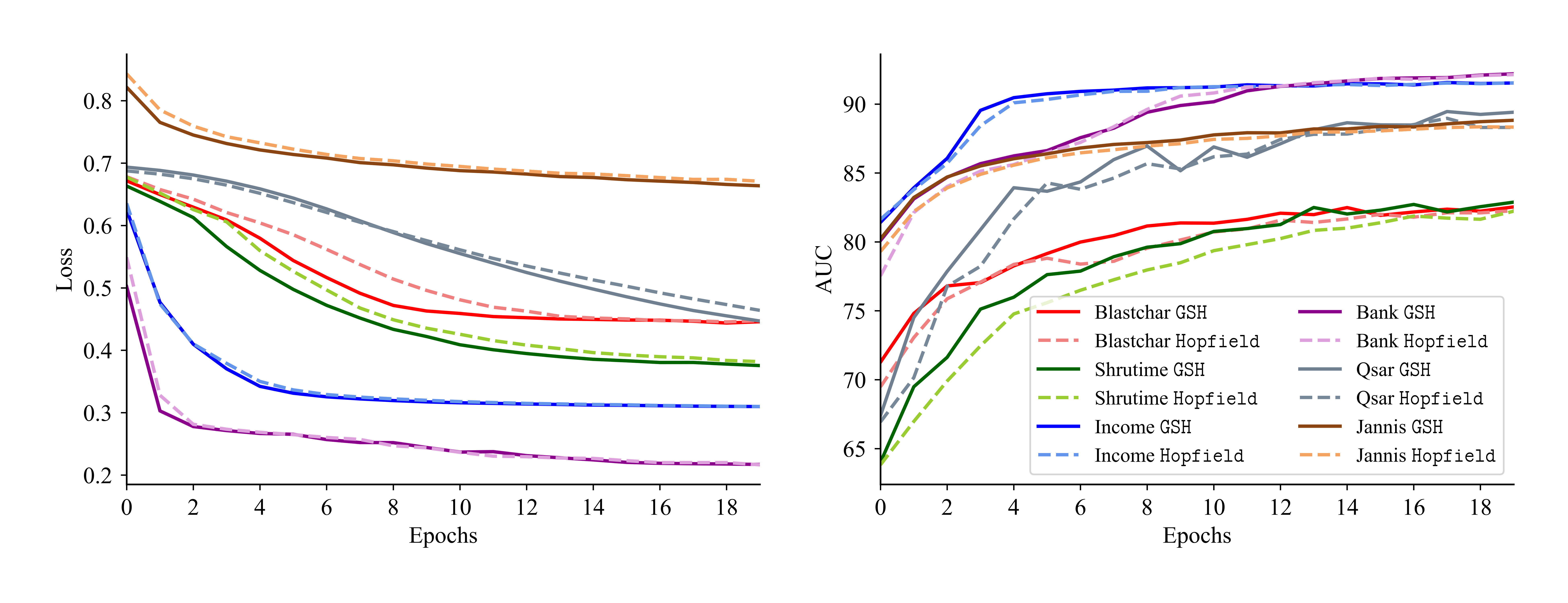}
    \caption{\small
    \textbf{Convergence Analysis.} We plot the validation loss and AUC score curves of the generalized sparse Hopfield model ($\mathtt{GSH}$) and the dense Hopfield model ($\mathtt{Hopfield}$). The results, as shown by the solid lines for $\mathtt{GSH}$, indicate that the sparse Hopfield model converges faster and yields superior accuracy compared to the dense Hopfield model.
    }
    \label{fig:feature}
\end{figure*}

\subsection{Convergence Analysis}
\label{sec:convergence}
We calculate the validation loss and AUC score using the same default parameters and compare them with the dense modern Hopfield model.
For ease of presentation, we plot the results of six datasets (Blastchar, Shrutime, Income, Bank, Qsar, and Jannis). 
We use the same hyperparameters for each dataset for both \texttt{GSH} and \texttt{Hopfield}. 
The results, averaged over 30 runs, are shown in \Cref{fig:feature}.
The results indicate that \texttt{GSH} converges faster and achieves an AUC score that is equal to or higher than \texttt{Hopfield}.

\subsection{Rotation Experiments}
\label{sec:rotation}
In \cref{tab:rot}, we conduct the following experiments on rotating the datasets and the BiSHopModule's direction, both individually and in combination:
\begin{enumerate}[leftmargin=2.3em]
\setlength\itemsep{0em}
    \item [(R1)] \textbf{Rotate the 2 Directions (Row-wise and Column-wise).}
    To validate the effectiveness of the bi-directional design in BiSHop, we conduct experiments by rotating these directions and reporting the performance and average results. The results indicate that the direction of BiSHop is vital for performance.
    \item [(R2)] \textbf{Rotate the Datasets.}
    Following the experimental setup in \cite{grinsztajn2022tree}, we randomly rotate datasets using a randomly generated special orthogonal matrix. The results indicate that BiSHop is robust against data rotation.
    \item [(R3)] \textbf{Rotate the 2 Directions and the Datasets.}
    To further validate our findings, we apply both (R1) and (R2). The results show a drop in performance across nearly every dataset and align with our findings in (R1) and (R2).
\end{enumerate}
The average AUC score across all datasets is reported for each type of rotation.

\begin{table*}[h]
\centering
\caption{
\textbf{
Comparing the Performance of Default BiSHop with Various Configurations on BiSHop Module and Datasets.}
We apply the following configurations to BiSHopModule and datasets to validate BiSHop's ability to tackle (\hyperref[item:C1]{C1}): rotate the 2 directions (R1), rotate the datasets (R2), and combined column-wise, row-wise, and rotate the 2 directions and the datasets (R3).
}
\vspace{0.5em}
\label{tab:rot}
\resizebox{ \textwidth}{!}{%
\begin{tabular}{lcccccccccc}
\toprule
Method/Dataset & Adult & Bank & Blastchar & Income & SeismicBumps & Shrutime & Spambase & Qsar & Jannis & Average \\
\midrule
BiSHop & 91.74 & 92.73 & 88.49 & 92.43 & 91.42 & 87.38 & 100 & 92.85 & 89.66 & \textbf{91.86} \\
(R1) & 91.52 & 92.04 & 88.38 & 91.69 & 89.81 & 85.83 & 100 & 93.65 & 86.55 & 91.05 \\
(R2) & 91.67 & 92.21 & 88.51 & 91.41 & 92.74 & 87.68 & 100 & 93.08 & 85.03 & 91.37 \\
(R3) & 91.44 & 92.07 & 85.68 & 91.62 & 89.92 & 85.85 & 100 & 94.18 & 87.1 & 90.88 \\
\bottomrule
\end{tabular}
}
\end{table*}

\subsection{Hierarchy of BiSHopModule}
\label{sec:stack}
In \cref{tab:stack}, we assess the impact of stacking different layers of BiSHopModule. We report the average AUC over all datasets for different layers of BiSHopModule.

\paragraph{Details.}
We progressively increase the layers within BiSHopModule from 1 to 8 in the Encoder and Decoder layers, keeping other parameters at their default settings. This approach allows us to examine how adding more layers affects the model's performance.

\paragraph{Results.}
\cref{tab:stack} summarizes the performance in AUC, averaged over all datasets for various layers. The results suggest that 4 layers are the optimal setting to maximize performance.

\begin{table*}[h]
\centering
\caption{
\textbf{Performance Comparison with Stacking Various Layers of BiSHopModule.}
We vary different layers of BiSHopModule in the encoder-decoder structure. The results suggest 4 layers of BiSHopModule may maximize the model performance.
}
\vspace{0.5em}
\label{tab:stack}
\resizebox{\textwidth}{!}{%
\begin{tabular}{lcccccccccc}
\toprule
Layer/Dataset & Adult & Bank & Blstchar & Income & SeismicBumps & Shrutime & Spambase & Qsar & Jannis & Average \\
\midrule
1 layer & 91.52 & 92.32 & 88.55 & 91.58 & 92.20 & 86.33 & 100 & 94.03 & 84.39 & 91.21 \\
2 layers & 91.56 & 92.21 & 88.71 & 91.66 & 90.83 & 87.5 & 100 & 93.77 & 82.84 & 91.00 \\
3 layers & 91.65 & 92.38 & 88.47 & 91.50 & 93.11 & 87.34 & 100 & 93.08 & 85.09 & 91.40 \\
4 layers & 91.58 & 92.28 & 88.54 & 91.47 & 92.98 & 87.24 & 100 & 93.52 & 85.40 & \textbf{91.45} \\
5 layers & 91.57 & 92.17 & 88.55 & 91.47 & 90.12 & 87.69 & 100 & 93.37 & 84.93 & 91.10 \\
6 layers & 91.65 & 92.26 & 88.48 & 91.46 & 92.47 & 85.05 & 100 & 91.14 & 85.11 & 90.85 \\
7 layers & 91.54 & 92.16 & 87.88 & 91.47 & 93.04 & 87.26 & 100 & 92.05 & 84.69 & 91.12 \\
8 layers & 91.56 & 91.96 & 88.09 & 91.54 & 93.04 & 82.98 & 100 & 93.88 & 84.71 & 90.86 \\
\bottomrule
\end{tabular}
}
\end{table*}

\subsection{Component Analysis for Regression Datasets}
We conduct ablation studies on 3 regression datasets to further validate the importance of various components in BiSHop. We randomly select 3 datasets from 'CR' in \ref{sec:exp_set} ‘Datasets II’ from our paper. For all experiments, we follow the same procedure as detailed in \ref{sec:ablation_comp}. We use R2 for evaluation.

\begin{table*}[h]
\centering
\caption{\textbf{Component Ablation}. We remove each component at one time and keep all other settings the same for regression datasets. For the experimental results, we prove that each component contributes to the model performance.}
\vspace{1em}
\resizebox{\textwidth}{!}{
\begin{tabular}{lccccccc}
\toprule
 Data & BiSHop & w/o Cat Emb & w/o Num Emb & w/o Patch Emb & w/o Decoder & w/o BiSHopModule \\ 
\midrule
361094 & 99.72 &	99.75 &	99.83 &	99.32 &	99.79 &	97.81 \\
361288 & 54.91 &	54.09 &	54.67 &	50.7 &	53.93 &	51.41 \\
361292	& 52.91 &	39.73 &	51.3 &	 52.76 &	51.56	 &34.14
\\
\midrule
Average  & 69.18  & 64.52 & 68.60 & 67.59 & 68.43  & 61.12 \\
\bottomrule
\end{tabular}
}
\label{tab:component_reg}
\end{table*}

\paragraph{Details.}
We remove different components from BiSHop for the randomly selected regression tasks from 'CR' in \ref{sec:exp_set} ‘Datasets II’ since 'CR' includes both categorical and numerical features. This ablation study is presented alongside the classification task analysis in \ref{sec:ablation_comp} to validate the importance of various component design philosophies in BiSHop.

\paragraph{Results.}
We summarize the R2 score in \cref{tab:component_reg}. Different components in BiSHop contribute to various degrees of performance enhancement. Notably, BiSHopModule and CatEmb have the most significant impact on regression tasks.

\clearpage
\section{Computational Time}

\paragraph{Computational Complexity.}
We summarize the computational complexity for each function used in BiSHop in \cref{tab:complexity}.
Here we use the same notation as introduced in the main paper:
$N^\text{cat}$ represents the number of categorical features;
$N^\text{num}$ represents the number of numerical features;
$N=N^\text{num}+N^\text{cat}$ represents the total number of features;
$G$ represents the embedding dimension;
$P$ represents the patch embedding dimension;
$D^\text{model}$ represents the hidden dimension;
$\text{len}(Q)$ represents the size of the query pattern;
$C$ represents the number of pooling vectors; and
$\text{len}(Y)$ represents the size of the memory pattern.
\begin{table*}[h]

\centering
\caption{Computational Complexity.}
\vspace{0.5em}
    \begin{tabular}{cc}

    \toprule
    Function Name & Time Complexity \\
    \midrule
        Categorical Embedding & $\mathcal{O}(G\times N^\text{cat} )$  \\
         Numerical Embedding & $\mathcal{O}(G\times N^\text{cat} )$ \\
         Patch Embedding & $\mathcal{O}(N \times P \times D^\text{model} )$ \\
         GSH & $\mathcal{O}(\text{len}(Y) \times \text{len}(Q) \times D^\text{model} )$\\
         GSHPooling &$\mathcal{O}(\text{len}(Q) \times C \times D^\text{model} \times P^2 )$ \\
         Merging & $\mathcal{O}(N \times P \times (D^\text{model})^2 )$\\
    \bottomrule
    \end{tabular}
     \label{tab:complexity}
\end{table*}

\paragraph{Computationally Time.}
For each dataset and hyperparameter configuration, the average training time for BiSHop varies from 30 minutes to 2 hours. 
Based on different hyperparameter settings, the number of our model parameters varies from $10^{7}$ to $10^{8}$.

\clearpage
\twocolumn
\def\arxivfont{\rm}
\bibliographystyle{plainnat}

\bibliography{refs}

\end{document}